%% file: main.tex
\documentclass[10pt,twocolumn,letterpaper]{article}

\usepackage[pagenumbers]{iccv} 

\usepackage[accsupp]{axessibility}  


\usepackage{amsmath}
\usepackage{amssymb}
\usepackage{mathtools}
\usepackage{amsthm}
\usepackage{mwe}
\usepackage{pifont}
\usepackage{wrapfig}
\usepackage{colortbl}
\usepackage{multirow}
\usepackage{multicol}
\usepackage{dsfont}

\DeclareMathOperator*{\argmin}{arg\,min}
\usepackage{makecell}
\usepackage{bbm}
\usepackage[T1]{fontenc}
\usepackage{algorithm}
\usepackage{algpseudocode}

\newcommand{\oursabb}{CMNet}

\definecolor{iccvblue}{rgb}{0.21,0.49,0.74}
\usepackage[pagebackref,breaklinks,colorlinks,allcolors=iccvblue]{hyperref}

\def\paperID{7257}
\def\confName{ICCV}
\def\confYear{2025}

\title{Combinative Matching for Geometric Shape Assembly}

\author{
Nahyuk Lee$^{1}$\thanks{Equal contribution} \quad\quad
Juhong Min$^{1,2}$\footnotemark[1] \quad\quad
Junhong Lee$^{1}$ \quad\quad
Chunghyun Park$^{1}$ \quad\quad
Minsu Cho$^{1,3}$ \vspace{0.15cm}\\
$^{1}$POSTECH \quad\quad\quad
$^{2}$Samsung Research America \quad\quad\quad 
$^{3}$RLWRLD\\
\href{https://nahyuklee.github.io/cmnet}{\texttt{\small https://nahyuklee.github.io/cmnet}}
}

\begin{document}

\maketitle
\input{sections/0_abstract}
\input{sections/1_introduction}
\input{sections/2_related_work}
\input{sections/3_method}
\input{sections/4_experiments}

\input{sections/5_conclusion}
\input{sections/6_acknowledgement}
{
    \small
    \bibliographystyle{ieeenat_fullname}
    \bibliography{egbib}
}
\input{sections/X_suppl}

\end{document}

%% file: sections/0_abstract.tex

\begin{abstract}
This paper introduces a new shape-matching methodology, \textbf{combinative matching}, to combine interlocking parts for geometric shape assembly. 
Previous methods for geometric assembly typically rely on aligning parts by finding identical surfaces between the parts as in conventional shape matching and registration. Specifically, we explicitly model two distinct properties of interlocking shapes: `identical surface shape' and `opposite volume occupancy.' 
Our method thus learns to establish correspondences across regions where their surface shapes appear identical but their volumes occupy the inverted space to each other. To facilitate this process, we also learn to align regions in rotation by estimating their shape orientations via equivariant neural networks. 
The proposed approach significantly reduces local ambiguities in matching and allows a robust combination of parts in assembly. 
Experimental results on geometric assembly benchmarks demonstrate the efficacy of our method, consistently outperforming the state of the art.
\end{abstract}

%% file: sections/1_introduction.tex

\section{Introduction}
\label{sec:introduction}

Geometric shape assembly, the task of reconstructing a target object from multiple fractured parts, plays a crucial role in diverse fields such as archaeology~\cite{mcbride2003archaeological, son2013axially, niu2022automatic}, medical imaging~\cite{zhang2021colde, li2023anatomy, marande2007mitochondrial}, robotics~\cite{thomas2018learning, harada2016proposal, zakka2020form2fit}, and industrial manufacturing~\cite{campi2022cad, canali2014automatic}.
Reliable assembly requires not only identifying common interfaces where parts align (\eg, mating surfaces) but also establishing robust feature correspondences that account for how different parts combine with each other.
This combinative process involves challenges in analyzing parts such as incomplete semantics, shape ambiguity, variations in orientation, and complexity in matching.

To address the challenges, prior work~\cite{chen2022neural, lee2024pmtr} has predominantly relied on aligning parts by finding identical surfaces between parts as in conventional shape matching and registration. 
The methods typically extract visual features and maximize similarity for positive matches at interfaces under the assumption of their high visual resemblance.
While technically sound, this approach often suffers from local ambiguities, where visually similar shapes from different parts are incorrectly matched, as shown in Fig.~\ref{fig:teaser}.
This conventional matching on pure shape similarity often results in incorrect matching and pose estimations, as it overlooks intrinsic properties between matching for registration and that for assembly\footnote{In this manuscript, we use term ``matching'' to denote {\em local} pairwise-compatibility test, reserving ``assembly'' for {\em global} placement of all parts.}.
This naturally raises the question: What do we miss in matching to address the challenges of geometric shape assembly?

\input{figures/teaser}

Drawing inspiration from construction and civil engineering, where male and female components combine to form stable structures, techniques such as mortise and tenon joints, tongue and groove connections, and dovetail joints~\cite{okamoto2021verification, matouvs2004analysis, kunecky2016mechanical} demonstrate how stability and precision are achieved not merely through visual resemblance but through combinative properties between parts.
Unlike surfaces designed to mirror each other, as in general scene/object alignment or registration tasks~\cite{qin2022geotransformer, huang2021predator}, the mating parts in geometric assembly~\cite{sellan2022breakingbad} are to be combined with each other, requiring attention to their mutual relationship.
Let us assume two corresponding points on the mating surfaces of two interlocking parts. The two points share identical surface shapes in their vicinities, but have opposite volume occupancy, \ie, the volume around one point occupies the inverted volume around the other point and vice versa. 
This observation reveals two distinct properties of interlocking shapes: {\em identical surface shape} and {\em opposite volume occupancy}. 
Reliable shape assembly thus needs to reflect both of the two properties in matching.  

To this end, we introduce a new shape-matching methodology for geometric assembly, dubbed {\em combinative matching}, which learns to match interlocking regions of parts. 
Unlike conventional matching for registration and assembly~\cite{huang2021predator, yu2021cofinet, qin2022geotransformer, lee2024pmtr, yao2024pare, yu2023rotation}, which commonly relies on shape similarity, combinative matching establishes correspondences across regions where their surface shapes appear identical but their volumes occupy the inverted space to each other.
Specifically, we train our model to learn: (1) shape orientations for consistent directional alignment, (2) surface shape descriptors for identical-shape matching, and (3) volume occupancy descriptors for inverted-volume matching.
These three objectives jointly help the model reduce local ambiguities, enhance its understanding of interlocking structures, and improve the overall accuracy of assembly.
Central to this approach is the use of equivariant and invariant descriptors, allowing both occupancy and shape descriptors to recognize orientation relationships through equivariance, while maintaining robustness to absolute pose through invariance.
Experimental results validate that our multi-faceted matching enables a robust, interlocking-aware geometric assembly, addressing the limitations of conventional matching.

%% file: figures/teaser.tex

\begin{figure}[t]
\begin{center}
\includegraphics[width=1\linewidth]{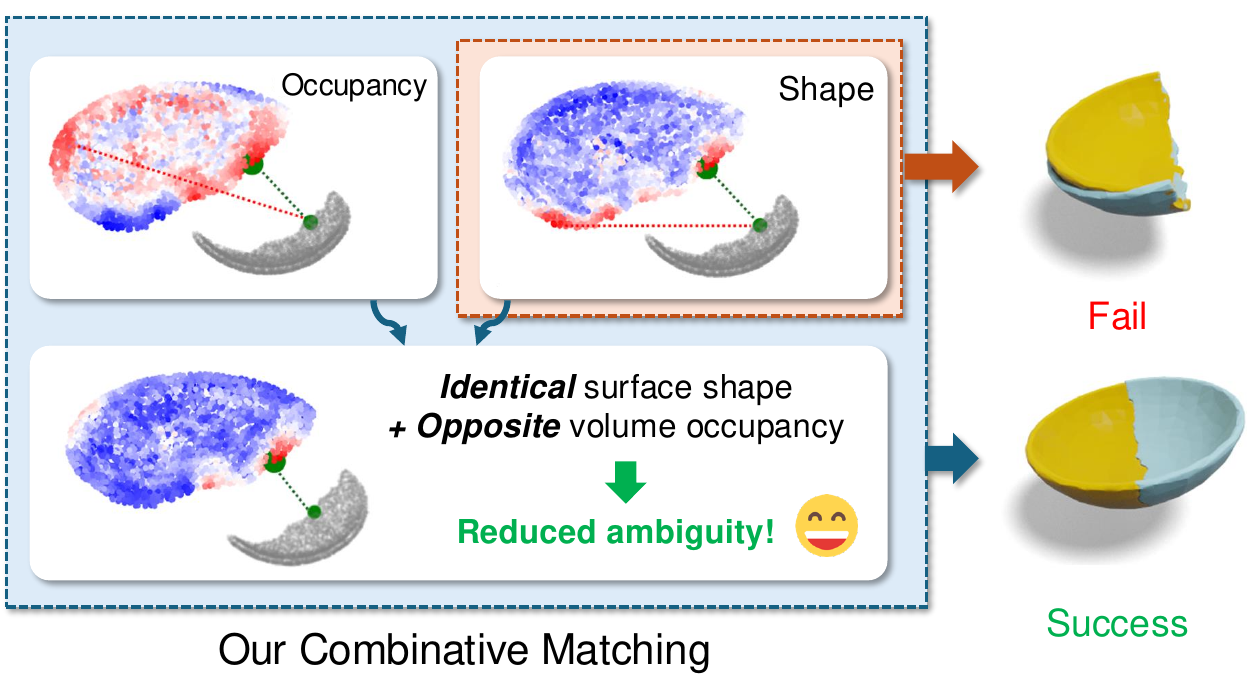}
\caption{
\textbf{Combinative matching.}
In contrast to conventional approaches to matching solely based on shape similarity, our combinative matching explicitly models two distinct properties of interlocking shapes, `identical surface shape' and `opposite volume occupancy,' and learns to establish correspondences across regions where their surface shapes appear identical but their volumes occupy the inverted space to each other. 
The figure shows the assembly of source (\textcolor[gray]{0.5}{gray}) and target (\textcolor{red}{red} \& \textcolor{blue}{blue}) parts, with a true match shown by green dots (\textcolor[HTML]{22741C}{$\bullet$}) connected by line. The color gradient on target points indicates correlation scores with the green source point, ranging from \textcolor{red}{red} (high) to \textcolor{blue}{blue} (low). Incorporating the volume occupancy (shown in this example), reduces visual ambiguities, achieving accurate assembly.
}
\label{fig:teaser}
\end{center}
\vspace{-6mm}
\end{figure}

%% file: sections/2_related_work.tex

\section{Related Work}
\label{sec:related_work}
\noindent\textbf{Shape assembly from parts.}
A common approach to reconstructing a target shape from its parts involves point cloud registration~\cite{huang2021predator, yu2021cofinet, qin2022geotransformer}, \ie, object or scene alignment tasks, which focus on localizing overlapping interfaces and establishing dense feature correspondences to predict alignment poses.
Shape assembly can be viewed as a challenging registration problem under extremely low-overlap conditions, \ie, surface overlap.
Existing assembly approaches can be broadly divided into two categories:
(1) first category includes methods that rely on direct pose regression using global embeddings for each part~\cite{schor2019componet, li2020global, wu2020lstm, huang2020dgl, li2024category, chen2022neural, wu2023leveraging}.
While efficient, these methods often lack fine-grained local detail, leading to inaccuracies.
Addressing this limitation, (2) methods such as Jigsaw~\cite{lu2023jigsaw} and PMTR~\cite{lee2024pmtr} employ dense feature matching to identify reliable correspondences, predicting poses based on the dense matches rather than direct regression, similar to those used in registration approaches~\cite{huang2021predator, yu2021cofinet, qin2022geotransformer}.
The dense matching methods~\cite{lu2023jigsaw, lee2024pmtr, huang2021predator, yu2021cofinet, qin2022geotransformer} are built on the assumption that mating interfaces exhibit high visual resemblance, leading to employ training objectives that maximize feature similarity for positive matches.
However, unlike general scene alignment, the assembly task requires more than resemblance-based matching alone:
Mating interfaces are shaped to interlock rather than mirror each other, demanding a deeper, context-aware understanding of structural complementarity beyond na\"ive feature similarity.

\smallbreak\noindent\textbf{Civil engineering and construction.}
A combinative design plays an essential role in creating durable assemblies as demonstrated by civil engineering techniques such as mortise and tenon joints~\cite{okamoto2021verification, liu2024investigating, ogawa2016theoretical, xie2021normal}, tongue and groove connections~\cite{matouvs2004analysis, canyurt2010strength, mollahassani2020dynamic}, dovetail joints~\cite{kunecky2016mechanical}, rabbet joints~\cite{yue2017robust, wang2020stiffness, zhou2022prediction}, and bridle joints~\cite{altinok2016mechanical}.
These methods share two key properties of combining parts: {\em surface resemblance} that ensures that mating parts align smoothly, and {\em volumetric complementarity} that reflects the design intention for parts to interlock in a structurally sound manner.   
Although existing shape assembly methods~\cite{lu2023jigsaw, lee2024pmtr} incorporate visual resemblance learning in their objectives, they typically lack the necessary learning to model the volumetric aspects of interlocking parts, which is essential for combinative matching for reliable assembly.

\smallbreak\noindent\textbf{Equivariance and invariance learning.}
Equivariance and invariance are essential properties in feature learning, especially for tasks involving spatial transformations, where understanding the relative pose relationship between parts is critical.
Equivariance ensures transformations applied to the input are reflected in the output, allowing models to retain key orientation information~\cite{qi2017pointnet, deng2021vector, ling2022vectoradam} and individual point orientations~\cite{luo2022equivariant, kim2023stable}, resulting in structure-aware representations. 
Invariant descriptors, on the other hand, are widely used for maintaining consistent feature representations regardless of transformations.
In geometric matching tasks, many efforts~\cite{yu2023rotation, yao2024pare, wang2022you, wang2023roreg} incorporate these descriptors to achieve rotation-invariant matching and alignment, demonstrating strong empirical performance.
Inspired by complementary geometric design in civil engineering, our study goes further beyond invariance-based simple visual matching, underscoring that capturing structural complementarity requires equivariant learning to enable models to understand the relative orientations of parts and their interdependency. Our experiments show that leveraging both equivariance and invariance enhances the model's ability to capture essential features for combinative matching.

\smallbreak\noindent\textbf{Complementary matching for assembly.} A recent work by Lu~\etal~\cite{lu2023jigsaw} presents the concept of a primal-dual descriptor to reflect viewpoint-dependent characteristics for surface matching. Similarly to our motivation, they intend to capture the essence of complementary geometry arising from fracture assembly. However, focusing on the characteristics of a local surface from two different directions, inward and outward, they separate a descriptor into primal and dual ones, and train them to align by intercrossing in matching, \ie, encouraging the primal descriptor of one part to resemble the dual descriptor of the other part in matching. 
Despite a similar motivation, the primal-dual matching method implements the geometric complementarity of mating parts simply by switching two distinct descriptors in matching, and train them to resemble in the primal-dual pair between mating parts; there is no clear distinction between the primal and dual descriptors in terms of their roles and effects.
This is clearly different from our approach that distinguishes the descriptor for surface shape, which is to be identical between mating parts, from that for volumetric occupancy, which is to be opposite between mating parts. 
As will be discussed in our experimental section and supplementary, the primal-dual descriptors fail to capture the interlocking properties of mating parts, and our combinative matching clearly outperforms in matching performance.  

%% file: sections/3_method.tex

\section{Proposed Approach}
\label{sec:proposed_approach}

\smallbreak
\noindent \textbf{Problem setup.}
Following previous shape assembly methods~\cite{lu2023jigsaw, lee2024pmtr, schor2019componet, li2020global, wu2020lstm, huang2020dgl, li2024category, chen2022neural, wu2023leveraging}, our method adopts a self-supervised learning approach:
Given a holistic target object, we decompose it into multiple parts, each represented as a point cloud and each undergoing a random rigid transformation.
The model takes this set of randomly transformed point clouds as input and predicts a corresponding set of transformation parameters, which are then applied to each part to reconstruct the original target object.
Model performance is evaluated by measuring the distances between the ground-truth and predicted assembly configurations, as well as the accuracy of transformation parameters.

\input{figures/combinative}

\input{figures/architecture}

\subsection{Combinative Matching}
\label{subsec:combinative_matching}
In this section, we introduce \textit{Combinative Matching}, a novel approach that addresses the dual requirements of geometric assembly: \emph{identical surface shape} and \emph{opposite volume occupancy}.
To ensure consistent assembly despite random transformations, our method first aligns local orientations between surface points, establishing a common reference frame.
Within this frame, shape descriptors align to match identical surface shapes, whereas occupancy descriptors are inversely aligned to ensure opposite volume occupancy, enabling parts to interlock properly (Fig.~\ref{fig:combinative}).

\smallbreak \noindent \textbf{Orientation alignment.}
For robust assembly, we require that surface points from different parts share a consistent orientation reference.
This alignment ensures that subsequent shape and occupancy features can be compared meaningfully.
To this end, we employ an equivariant network $f_{\text{d}}$, which takes a point cloud $\mathbf{P} \in \mathbb{R}^{N \times 3}$ or $\mathbf{Q} \in \mathbb{R}^{M \times 3}$ and predicts orientations $\mathbf{F}^{\text{P}}_{\text{d}} = f_{\text{d}}(\mathbf{P}) \in \mathbb{R}^{N \times 3 \times 3}$ and $\mathbf{F}^{\text{Q}}_{\text{d}} = f_{\text{d}}(\mathbf{Q}) \in \mathbb{R}^{M \times 3 \times 3}$ with $(\mathbf{F}^{\text{Q}}_{\text{d}})_{i}, (\mathbf{F}^{\text{P}}_{\text{d}})_{i} \in \text{SO(3)}$, where $N$ and $M$ are the numbers of sampled points for the respective parts.
The training loss for orientation alignment is defined as the difference between aligned orientations:
\begin{align}
    \label{eq:orientation_loss}
    \mathcal{L}_{\text{d}} 
    = \frac{1}{|\mathcal{C}|} \sum_{(i,j) \in \mathcal{C}} 
    \bigl\| (\mathbf{F}^{\text{P}}_{\text{d}})_{i} \,\mathbf{R}^{\text{P}} \;-\; (\mathbf{F}^{\text{Q}}_{\text{d}})_{j} \,\mathbf{R}^{\text{Q}} \bigr\|_{F},
\end{align}
where $\mathcal{C}$ is the set of indices for positive matches, $\mathbf{R}^{\text{P}}$ and $\mathbf{R}^{\text{Q}}$ are the ground-truth rotations of parts $\mathbf{P}$ and $\mathbf{Q}$, and $\|\cdot\|_{F}$ is the Frobenius norm.
By minimizing this loss, the network learns to predict orientations that can be used to extract rigid transformation-invariant occupancy and shape descriptors in the subsequent matching steps, enabling stable assembly regardless of initial part positions.

\smallbreak \noindent \textbf{Surface shape matching.}
For identical-shape matching, we require shape descriptors that capture \emph{identical} surface characteristics.
For this purpose, given the learned surface shape embeddings $\mathbf{F}_{\text{s}}^{\text{P}} = f_{\text{s}}(\mathbf{P}) \in \mathbb{R}^{N \times d_{\text{s}}}$ and $\mathbf{F}_{\text{s}}^{\text{Q}} = f_{\text{s}}(\mathbf{Q}) \in \mathbb{R}^{M \times d_{\text{s}}}$ and a set of all indices for all points on the mating surface $\mathcal{I}$, we employ the standard circle loss~\cite{sun2020circle} without modification as follows:
\begin{align}
    \label{eq:shape_loss}
    \mathcal{L}_{\text{s}} 
    = \underset{i \sim \mathcal{I}}{\mathbb{E}} \left[ 
        \log \!\Bigl( 
          \sum_{j \in \mathcal{E}_{\text{p}}(i)} 
          e^{\alpha_{ij}(d_{ij}^{\text{p}} - \Delta_{\text{p}})}
          \cdot
          \sum_{k \in \mathcal{E}_{\text{n}}(i)} 
          e^{\beta_{ik}(\Delta_{\text{n}} - d_{ik}^{\text{n}})}
        \Bigr) 
    \right],
\end{align}
where $d_{ij}^{\text{p}} = \|\hat{\mathbf{F}}_{\text{s},i}^{\text{P}} - \hat{\mathbf{F}}_{\text{s},j}^{\text{Q}}\|_2$ represents the L2 distances between shape features in the embedding space, where $\hat{\mathbf{F}}$ denotes the L2-normalized features, and $\mathcal{E}_{\text{p}}(i), \mathcal{E}_{\text{n}}(i)$ are positive/negative correspondences for index $i$, and $\Delta_{\text{p}}, \Delta_{\text{n}}$ are margin hyperparameters.
The positive and negative weights are computed as $\alpha_{ij}=\gamma[d_{ij}^{\text{p}}-\Delta_\text{p}]_{+}$ and $\beta_{ik}=\gamma[\Delta_\text{n}-d_{ik}^{\text{n}}]_{+}$, with scale factor $\gamma$.
This formulation encourages the distance for positive matches to fall below the threshold $\Delta_{\text{p}}$ while pushing the distance for negative ones to exceed $\Delta_{\text{n}}$, similar to conventional approaches that identify identical surface geometries.

\smallbreak \noindent \textbf{Volume occupancy matching.}
A key insight for interlocking parts is that their local volumes must \emph{occupy} opposite spaces at the interface.
Specifically, if one region is occupied, the corresponding region in the mating part should be unoccupied, and vice versa—creating the interlocking relationship necessary for proper assembly.
We encode this idea by learning volume occupancy descriptors $\mathbf{F}_{\text{o}}^{\text{P}} = f_{\text{o}}(\mathbf{P}) \in \mathbb{R}^{N \times d_{\text{o}}}$ and $\mathbf{F}_{\text{o}}^{\text{Q}} = f_{\text{o}}(\mathbf{Q}) \in \mathbb{R}^{M \times d_{\text{o}}}$ that are invariant to rigid transformations through the orientation alignment.
To ensure that occupancy descriptors from corresponding regions have opposite values, we define the occupancy matching loss using a variant of circle loss~\cite{sun2020circle}:
\begin{align}
    \label{eq:occupancy_loss}
    \mathcal{L}_{\text{o}} 
    = \underset{i \sim \mathcal{I}}{\mathbb{E}} \left[ 
        \log \!\Bigl( 
          \sum_{j \in \mathcal{E}_{\text{p}}(i)} 
          e^{\alpha_{ij}(s_{ij}^{\text{p}} - \Delta_{\text{p}})}
          \cdot
          \sum_{k \in \mathcal{E}_{\text{n}}(i)} 
          e^{\beta_{ik}(\Delta_{\text{n}} - s_{ik}^{\text{n}})}
        \Bigr) 
    \right],
\end{align}
where $s_{ij}^{\text{p}} = \|\hat{\mathbf{F}}_{\text{o},i}^{\text{P}} +\hat{\mathbf{F}}_{\text{o},j}^{\text{Q}}\|_2 \approx \cos(\mathbf{F}_{\text{o},i}^{\text{P}}, \mathbf{F}_{\text{o},j}^{\text{Q}})$ represents the cosine similarity measures in the occupancy embedding space, and the positive and negative weights are computed as $\alpha_{ij}=\gamma[s_{ij}^{\text{p}}-\Delta_\text{p}]_{+}$ and $\beta_{ik}=\gamma[\Delta_\text{n}-s_{ik}^{\text{n}}]_{+}$.
It is worth noting that while conventional circle loss typically uses distance metrics to bring positive pairs closer together, our approach leverages cosine similarity to explicitly encourage \emph{opposite} occupancy between positive pairs. 
This approach treats occupied-unoccupied pairs as \emph{positive} matches, encouraging their descriptors to be opposite, while penalizing non-matching pairs.
Thus, the model learns to identify complementary volumes that interlock, rather than matching identical geometries.

Consequently, our Combinative Matching effectively achieves the two essential desiderata for geometric shape assembly: matching identical surface shapes at interfaces and ensuring opposite volume occupancy for proper interlocking, invariant to initial part orientations.

\vspace{4mm}
\subsection{Combinative Matching Network}
\label{sec:network}
We now present the proposed framework that achieves combinative matching through the proposed objectives, capturing the multi-faceted aspects essential for assembly: orientation, shape, and occupancy.
Figure~\ref{fig:architecture} illustrates the overall architecture, which consists of five parts: (a) feature extraction and orientation alignment, (b) surface shape matching, (c) volume occupancy matching, (d) transformation estimation, and (e) training objective.

\smallbreak
\noindent \textbf{(a) Feature extraction and orientation alignment.}
For effective combinative matching, surface shape descriptors should ideally be rotation-invariant to ensure robustness across various orientations, while volume occupancy descriptors must retain direction-aligned information within their embedding space to enable complementary alignment.
Therefore, prior to applying shape and occupancy matching, we require rotation-invariant features that also encode orientation-consistent information.

To address these requirements, we design a feature embedding network that can embed both clues of orientation-consistency and invariance into a unified representation.
We employ VN-EdgeConvs~\cite{deng2021vector}, which takes as input a pair of point clouds $\mathbf{P}$ and $\mathbf{Q}$ and provides rotation-equivariant features $\mathbf{F}^{\text{P}}_{\text{eqv}}, \mathbf{F}^{\text{Q}}_{\text{eqv}} \in \mathbb{R}^{{K} \times D \times 3}$ for each input where ${K}$ corresponds to $N$ or $M$ depending on the input point cloud, $\mathbf{P}$ and $\mathbf{Q}$, respectively.
Next, an orientation hypothesizer, implemented with VN-Linear~\cite{deng2021vector}, processes the equivariant features, followed by Gram-Schmidt process with cross-product operation to provide orientations for each point, denoted as $\mathbf{F}^{\text{P}}_{\text{d}}, \mathbf{F}^{\text{Q}}_{\text{d}} \in \mathbb{R}^{{K} \times 3 \times 3}$.
We obtain rotation-invariant features by taking the dot-product between the equivariant features and the orientation matrices: $\mathbf{F}^{\text{P}}_{\text{inv}} = \mathbf{F}^{\text{P}}_{\text{eqv}} \cdot \mathbf{F}^{\text{P}\top}_{\text{d}}$\footnote{We refer the readers to the supplementary for a detailed proof.}, with the same calculation for $\mathbf{F}^{\text{Q}}_{\text{inv}}$.
To ensure these invariant features are aligned consistently with orientation information, we employ the orientation training objective $\mathcal{L}_{\text{d}}$ from Eq.~\ref{eq:orientation_loss}, which encourages the features to encode orientation-aligned information while maintaining rotation-invariance, thus ensuring complementary alignment and visual consistency for both shape and occupancy descriptor learning.

\smallbreak
\noindent \textbf{(b) Surface shape matching branch.}
Similar to the way mating surfaces of male and female parts exhibit compatible appearance, we require a distinct feature representation that effectively encodes appearance information to learn shape compatibility.
To achieve this, we introduce a dedicated branch that takes the rotation-invariant features $\mathbf{F}^{\text{P}}_{\text{inv}}, \mathbf{F}^{\text{Q}}_{\text{inv}}$ to embed them into surface shape descriptors $\mathbf{F}^{\text{P}}_{\text{s}}, \mathbf{F}^{\text{Q}}_{\text{s}} \in \mathbb{R}^{K \times d_{\text{s}}}$, using a three-layer MLP, followed by LeakyReLU.
Applying the surface shape matching objective $\mathcal{L}_{\text{s}}$ from Eq.~\ref{eq:shape_loss} ensures that matching surfaces with similar appearance are correctly aligned, forming a reliable basis for the subsequent transformation estimation for assembly.

\smallbreak
\noindent \textbf{(c) Volume occupancy matching branch.}
To capture occupancy, we introduce another dedicated branch to learning occupancy descriptors, allowing the model to recognize complementary alignment requirements.
This branch begins by taking the $\mathbf{F}^{\text{P}}_{\text{inv}}, \mathbf{F}^{\text{Q}}_{\text{inv}}$ which encode orientation-consistency information in their representations and provide occupancy descriptors $\mathbf{F}^{\text{P}}_{\text{o}}, \mathbf{F}^{\text{Q}}_{\text{o}} \in \mathbb{R}^{K \times d_{\text{o}}}$, using a three-layer MLP with parameters distinct from those in shape matching branch, followed by a Tanh activation.
To enforce correct alignment of complementary surfaces, we apply the volume occupancy matching objective $\mathcal{L}_{\text{o}}$ from Eq.~\ref{eq:occupancy_loss} which penalizes similarity between descriptors of complementary (occupied {\em vs.} unoccupied) regions, thereby ensuring corresponding surfaces interlock stably.

\smallbreak
\noindent \textbf{(d) Transformation estimation.}
With the surface shape and volume occupancy descriptor pairs obtained, we construct a cost matrix $\mathbf{C}\in\mathbb{R}^{N \times M}$ that encodes unified correlations across both shape and occupancy characteristics:
\begin{align}
    \mathbf{C} = (\mathbf{F}^{\text{P}}_{\text{s}} \cdot \mathbf{F}^{\text{Q}\top}_{\text{s}} - \mathbf{F}^{\text{P}}_{\text{o}} \cdot \mathbf{F}^{\text{Q}\top}_{\text{o}}) / Z,
\end{align}
where $Z$ is a normalization constant.
In this formulation, the shape descriptors are learned to be similar for positive matches, meaning that their dot product reflects a direct measure of `similarity' while the occupancy descriptors are trained with the opposite objective, implying their dot product instead represents the `dissimilarity'.
By negating the dissimilarity, $\mathbf{C}$ becomes a similarity measure, integrating the visual similarity with the volumetric complementarity, forming a comprehensive, {\em combinative} cost matrix.

To obtain a reliable set of correspondence indices $\hat{\mathcal{C}}$, we first apply an Optimal Transport (OT) layer~\cite{sarlin2020superglue} to encourage one-to-one correspondence, then collect the top-$k$ correspondences ($k=128$), resulting in $|\hat{\mathcal{C}}| = 128$.
Finally, we estimate the transformation between the pair of point clouds using weighted SVD~\cite{choy2020deep}, formulated as follows:
\vspace{2mm}
\begin{align}
\mathbf{R}^{*}, \mathbf{t}^{*} = \argmin_{\mathbf{R},\mathbf{t}} \sum_{(i, j) \in \hat{\mathcal{C}}} w_{ij} \|\mathbf{R} \mathbf{P}_i +\mathbf{t}-\mathbf{Q}_j \|^2_2,
\end{align}
where $w_{ij}$ represents the weight, \eg, the output of OT, for match $(i,j)$.
For multi-part assembly, we adopt the same transformation estimation method as used in~\cite{lee2024pmtr}, of which implementation details are provided in the supplementary.

\smallbreak
\noindent \textbf{(e) Training objective.}
Following previous point cloud matching methods~\cite{qin2022geotransformer, lee2024pmtr, yao2024pare}, we incorporate a point matching loss $\mathcal{L}_{\text{p}}$~\cite{qin2022geotransformer}, cross-entropy loss between ground-truth and predicted match probabilities.
By integrating orientation, shape, and occupancy losses along with point matching loss, the final training objective is formulated as:
\begin{align}
    \mathcal{L} = \lambda_{\text{d}} \mathcal{L}_{\text{d}} + \lambda_{\text{s}} \mathcal{L}_{\text{s}} + \lambda_{\text{o}} \mathcal{L}_{\text{o}} + \mathcal{L}_{\text{p}},
\end{align}
where $\lambda_{\text{d}} = 0.1$, $\lambda_{\text{s}} = 0.5$, and $\lambda_{\text{o}} = 0.5$ are weighting coefficients, balancing contributions of different matchings.

%% file: figures/combinative.tex

\begin{figure}[t]
\begin{center}
\includegraphics[width=\linewidth]{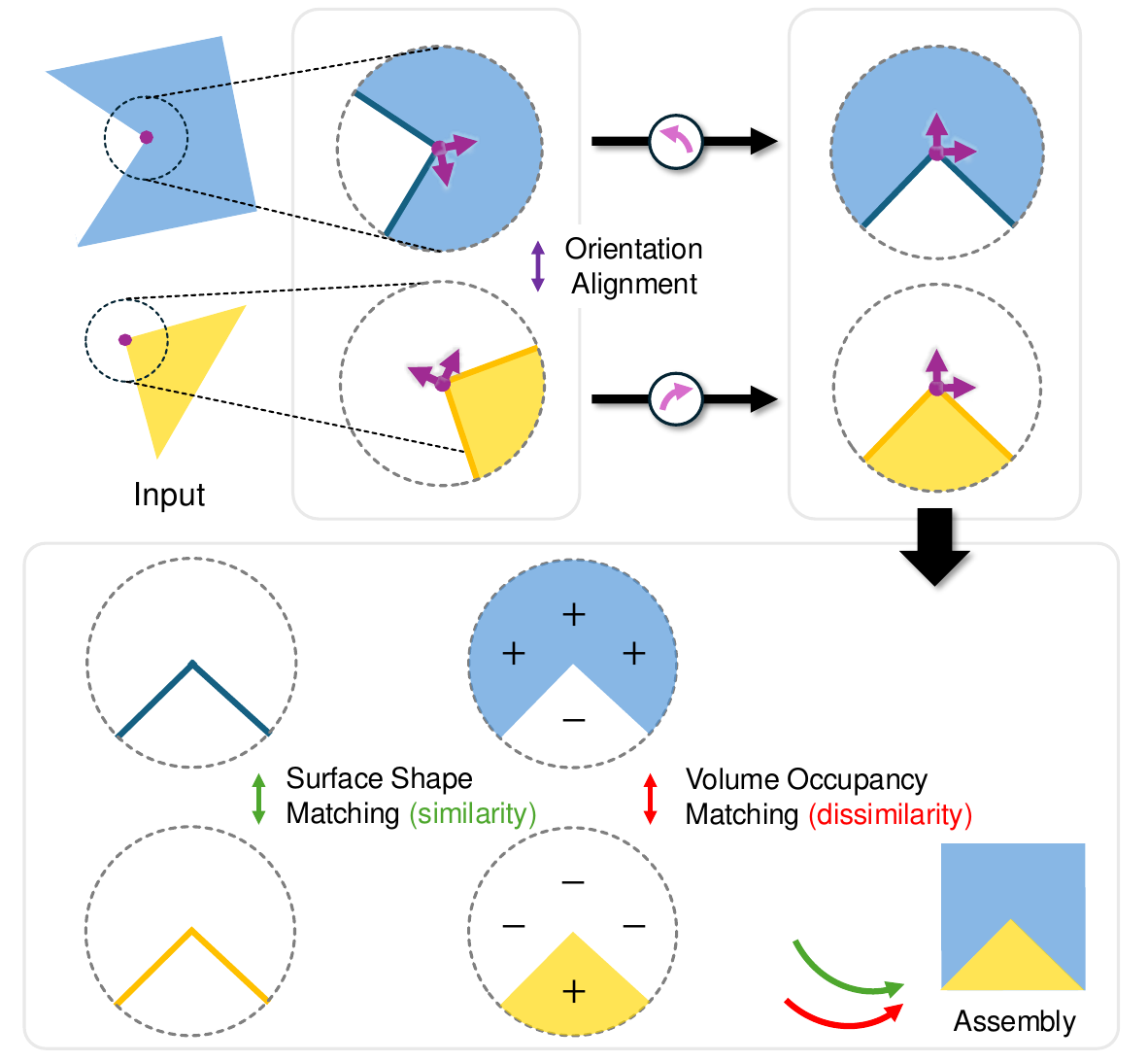}
\vspace{-6mm}
\caption{Main concept of our combinative matching.}
\label{fig:combinative}
\end{center}
\vspace{-8mm}
\end{figure}

%% file: figures/architecture.tex

\begin{figure*}[!t]
\begin{center}
\includegraphics[width=1\linewidth]{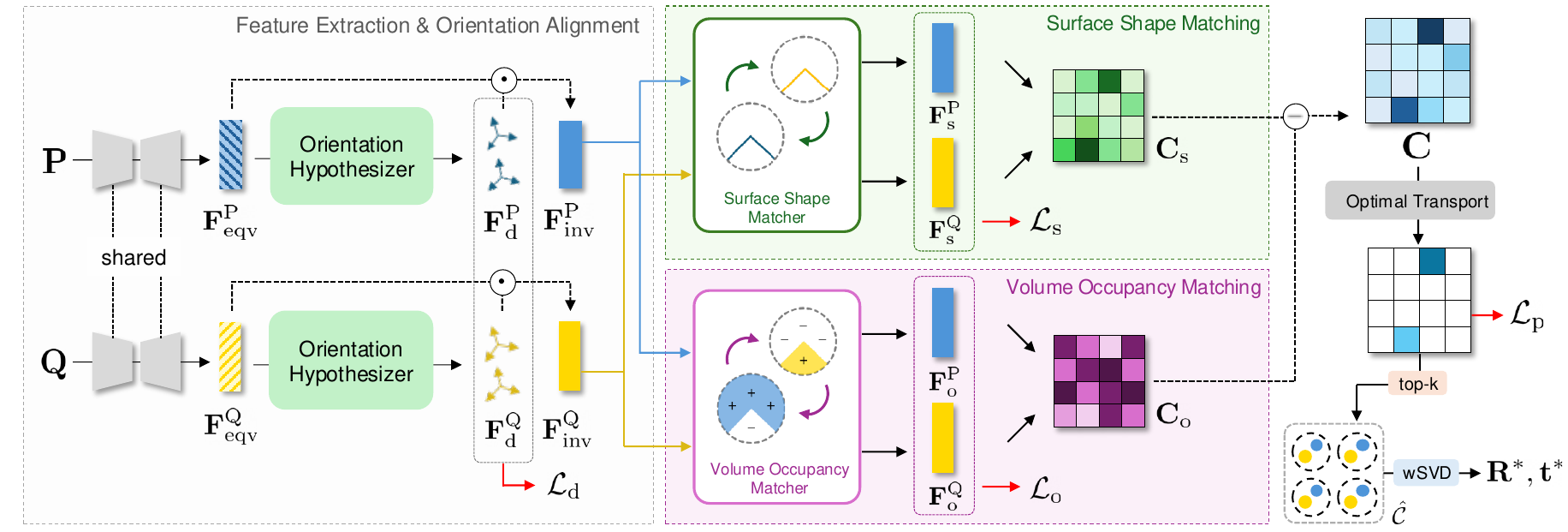}
\caption{\textbf{Overall architecture.} 
Here, we show core components of (a) feature embedding network, (b) surface shape matching branch, (c) volume occupancy matching branch, and (d) transformation estimation. 
We refer the readers to Sec.~\ref{sec:network} for details of each component.}
\label{fig:architecture}
\end{center}
\vspace{-4mm}
\end{figure*}

%% file: sections/4_experiments.tex

\section{Experiments}
\label{sec:experiments}

\subsection{Dataset and Evaluation Metrics}
\label{sec:dataset}

\smallbreak\noindent\textbf{Dataset.}
For our experiments, we use the large-scale, standard geometric assembly dataset, Breaking Bad~\cite{sellan2022breakingbad}, consisting of multiple fractured parts of target objects, categorized into two main subsets: \texttt{everyday} and \texttt{artifact}. 
For pairwise shape assembly, we focus specifically on its 2-part subset, while we utilize the entire dataset for multi-part assembly, which consists of objects with 2 to 20 parts.
Our experiments are conducted on the volume-constrained Breaking Bad dataset in which the volume of every piece is at least 1/40 of the total shape volume, reducing extreme point density imbalance.
For vanilla Breaking Bad benchmark evaluation results, we refer to the supplementary.

\input{figures/orientation}

\input{figures/heatmap}

\smallbreak\noindent\textbf{Evaluation Metrics.}
We use the evaluation metrics used in PMTR~\cite{lee2024pmtr} to validate our method:
(1) RMSE between ground truth and predicted rotation and translation parameters, (2) CoRrespondence Distance (CRD), the average distance between positive matches on mating surfaces, and (3) Chamfer Distance (CD) between the input and model-predicted assemblies.
Note that CRD provides a more reliable measure than RMSE(R,T) as CRD directly assesses assembly quality while RMSE(R,T) measures relative pose differences without explicitly considering alignment accuracy.
Following Lee~\etal~\cite{lee2024pmtr}, our evaluation is performed with \textit{relative poses} between parts instead of absolute ones to solely focus on the assembly, not absolute positioning.

\subsection{Implementation Details}
\label{sec:implementation}
We implement our model using PyTorch Lightning~\cite{falcon2019lightning}. 
All experiments were conducted on 4 NVIDIA GeForce RTX 3090 GPUs. 
We utilize the AdamW~\cite{loshchilovdecoupled} optimizer with an initial learning rate of $1 \times 10^{-2}$, employing a cosine scheduler set for 90 and 120 epochs on the respective \texttt{everyday} and
\texttt{artifact} subsets.
Following previous work of~\cite{lu2023jigsaw, lee2024pmtr}, we uniform-sample approximately 5,000 points on the surface per holistic object, with each part allocated a subset of points proportional to its surface area.

\subsection{Experimental Results and Analyses}

\smallbreak\noindent\textbf{Analysis on learned orientation} $\mathbf{F}^{\text{P}}_{\text{d}}, \mathbf{F}^{\text{Q}}_{\text{d}}$\textbf{.}
We begin by analyzing the learned orientations $(\mathbf{F}^{\text{P}}_{\text{d}})_{i}, (\mathbf{F}^{\text{Q}}_{\text{d}})_{i} \in \text{SO(3)}$ to observe the types of information captured through combinative matching, such as the directionality of occupied/unoccupied regions, magnitudes of local concavity or convexity, surface normal, and other relevant geometric properties.
For the analysis, we represent each orientation $(\mathbf{F}_{\text{d}})_{i} = [\mathbf{x}_i, \mathbf{y}_i, \mathbf{z}_i]$ for all $i$, where $\mathbf{x}_i, \mathbf{y}_i \in \mathbb{R}^{3}$ are orthonormal vectors, and $\mathbf{z}_i$ given by $\mathbf{x}_i \times \mathbf{y}_i$.
We focus on visualizing the scaled\footnote{They are scaled by the magnitudes of the input vectors for the Gram-Schmidt process, specifically for analysis purposes.} vectors $\mathbf{x}_i$ and $\mathbf{y}_i$, omitting $\mathbf{z}_i$ as it is redundant for interpretative purposes.

Figure~\ref{fig:orientation} visualizes the vectors $\{\mathbf{x}_i\}_{i \in \mathcal{I}}$ and $\{\mathbf{y}_i\}_{i \in \mathcal{I}}$ for both the source $\mathbf{F}^{\text{P}}_{\text{d}}$ and target $\mathbf{F}^{\text{Q}}_{\text{d}}$.
Through this visualization, we observe several notable patterns:
For both $\mathbf{x}_i$ and $\mathbf{y}_i$, (1) source and target orientations are aligned in parallel, as enforced by our training objective $\mathcal{L}_{\text{d}}$ (Eq.~\ref{eq:orientation_loss}).
For $\mathbf{x}_i$, we observe that (2) The learned $\mathbf{x}_i$ vectors are consistently directed toward the center of the mating surface, (3) staying parallel to the 2D plane of the mating surface lies, indicating our model has learned a stable orientation that respects the geometry of mating surfaces.
For $\mathbf{y}_i$, we observe that:
(4) vectors on convex regions (where the surface extends outward into occupied space) point outward, while those on concave regions (where the surface recedes) point inward.
(5) The magnitudes of $\mathbf{y}_i$ correlate with the degree of convexity or concavity at each point, indicating an awareness of surface curvature.
These results imply that the learned orientations not only differentiate between convex and concave structures but also capture complementarity and directional alignment {\em without any explicit supervisions dedicated to these aspects from (2) to (5)}, highlighting the efficacy of the proposed combinative matching in intuitive learning of integral properties for `combining' elements.

\input{figures/feat_tsne}

\smallbreak\noindent\textbf{Analysis on learned correlations.}
To validate how the proposed combinative matching resolves a limitation of conventional shape-based matching (\eg, local ambiguity), we compare correlation matrices for shape, occupancy, and the combined cost matrix:
specifically, $\mathbf{C}_{\text{s}} = \mathbf{F}^{\text{P}}_{\text{s}}\mathbf{F}^{\text{Q}\top}_{\text{s}} \in \mathbb{R}^{N \times M}$, $\mathbf{C}_{\text{o}} = \mathbf{F}^{\text{P}}_{\text{o}}\mathbf{F}^{\text{Q}\top}_{\text{o}} \in \mathbb{R}^{N \times M}$, and $\mathbf{C}$.
For this analysis, we select a single point on the source's mating surface (index $i$) and examine its similarity distributions of these correlations: $(\mathbf{C}_{\text{s}})_i, (\mathbf{C}_{\text{o}})_i, \mathbf{C}_i \in \mathbb{R}^{M}$.
These distributions are visualized as heatmaps, with red and blue colors indicating high and low similarities, respectively (we invert the color for $\mathbf{C}_{\text{o}}$ to reflect its representation of dissimilarity).
Figure~\ref{fig:heatmap} presents the visualized distributions.

Based solely on the surface shape distribution, the best target match for the $i$-th source point is located in a broad area due to the similar appearance of surrounding points, resulting in local ambiguity.
The volume occupancy distribution, on the other hand, shows large scores are nearly uniformly spread across the surface, with a slightly higher score near the true match, indicating it provides complementary information yet lacks distinct localization.
By combining shape and occupancy information, the local ambiguity and match confidence uncertainty are resolved;
the score at the true match is significantly higher, enabling precise alignment of the source point with its correct match on the target, verifying that the combinative matching effectively enhances precision by resolving the local ambiguity.

\input{tables/3_abl_combined}
\input{tables/suppl_2_generalization}

\smallbreak\noindent\textbf{Ablation studies.}
To assess the contributions of key components in our method, we conduct ablation studies on \texttt{everyday} subset.
First, Tab.~\ref{table:res_real_data_wide} (a) examines the choice of affinity metric ($s^{\text{p}}$ and $s^{\text{n}}$ in Eq.~\ref{eq:occupancy_loss}) and the impact of orientation loss $\mathcal{L}_{\text{o}}$.
Using L2 distance instead of cosine similarity, or omitting orientation loss during training, results in consistent performance drops, implying the importance of both complementarity and orientation learning in assembly.
Second, Tab.~\ref{table:res_real_data_wide} (b) evaluates the impact of equivariant network~\cite{deng2021vector} and surface shape \& volume occupancy matching branches.
When the equivariant backbone is replaced with a standard point embedding network, \eg, DGCNN~\cite{wang2019dynamic}, we observe a substantial drop in CRD, verifying the importance of learning orientation-awareness and rotation-invariance in assembly.
Consistent accuracy drops in the absence of either shape or occupancy matching branch demonstrate that both branches work synergistically to enhance alignment.

\smallbreak\noindent\textbf{Learned descriptor analysis.}
In the proposed network, $\mathbf{F}_{\text{s}}$ and $\mathbf{F}_{\text{o}}$ are optimized through the \textbf{{\em opposing}} objectives:
$\mathcal{L}_{\text{s}}$ clusters mating surface features for positive matches while separating negative ones, whereas $\mathcal{L}_{\text{o}}$ penalizes features for positive matches, each within its respective embedding space.
To examine their clustering behavior, we project the invariant features $\mathbf{F}_{\text{inv}}$, shape descriptors $\mathbf{F}_{\text{s}}$, and occupancy descriptors $\mathbf{F}_{\text{o}}$ into a 2D space using t-SNE and visualize the results in Fig.~\ref{fig:tsne}.

For invariant features of mating surface, we observe neither separation nor adhesion in their embedding space, implying that the invariance property alone does not provide significant feature distinction.
In contrast, the shape descriptors lying on mating surface are tightly clustered as enforced by $\mathcal{L}_{\text{s}}$, supported by the visual resemblance of the interface.
Meanwhile, the occupancy descriptors on mating surface are more widely dispersed, guided by $\mathcal{L}_{\text{o}}$.
The results collectively highlight the efficacy of proposed learning objectives in shaping the embedding space, reflecting both visual and volumetric properties of mating surfaces.

\input{tables/1_pairwise_bbad}

\input{figures/pairwise_qual}

\subsection{Model generalizability}
To demonstrate the generalizability of our approach, we conduct transferability experiments within the Breaking Bad dataset~\cite{sellan2022breakingbad}.
Specifically, we evaluate our model, trained on the \texttt{everyday} subset, on the \texttt{artifact} subset, and vice versa.
The \texttt{everyday} subset primarily consists of common objects relevant to computer vision and robotics applications, whereas the \texttt{artifact} subset focuses on archaeological objects, representing a notable domain shift between the two subsets.

Table~\ref{table:suppl_generalization} summarizes the transferability results.
The proposed method consistently outperforms state-of-the-art baselines, achieving higher CRDs across cross-subset evaluations.
This highlights the robustness of our model in adapting to different data domains, underscoring its efficacy in capturing task-oriented features of orientation, shape, and occupancy generalizable across distinct object categories.

\input{tables/2_mpa_bbad}
\input{figures/mpa_qual}

\subsection{Comparison with State of the Arts}
To validate the efficacy of the proposed method, we compare it with recent baselines on pairwise assembly in Tab.~\ref{table:pairwise_v3} and Fig.~\ref{fig:pairwise_qual}.
In terms of CRD and CD, our method outperforms all the baselines in both \texttt{everyday} and \texttt{artifact} subsets, achieving relative CRD improvements of 28\% and 18\%, respectively, over the previous state of the art~\cite{lee2024pmtr}.

We further evaluate our method on multi-part assembly\footnote{As noted in Sec.~\ref{sec:dataset}, our comparisons are conducted on the volume-constrained Breaking Bad dataset. The evaluation results on the vanilla Breaking Bad dataset are provided in our supplementary material.} with additional metrics, Part Accuracy (PA)~\cite{li2020learning, lee2024pmtr}, and compare the results in Tab.~\ref{table:mpa} and Fig.~\ref{fig:mpa_qual}, where ours consistently shows superior numbers compared to baselines.
The results confirm that, unlike methods that rely solely on visual cues~\cite{qin2022geotransformer,lee2024pmtr}, our combinative matching enables more reliable shape assembly, as evident from Figs.~\ref{fig:pairwise_qual} and~\ref{fig:mpa_qual}.
We refer to the supplementary for additional qualitative results.

%% file: figures/orientation.tex

\begin{figure}[t]
\begin{center}
\includegraphics[width=.95\linewidth]{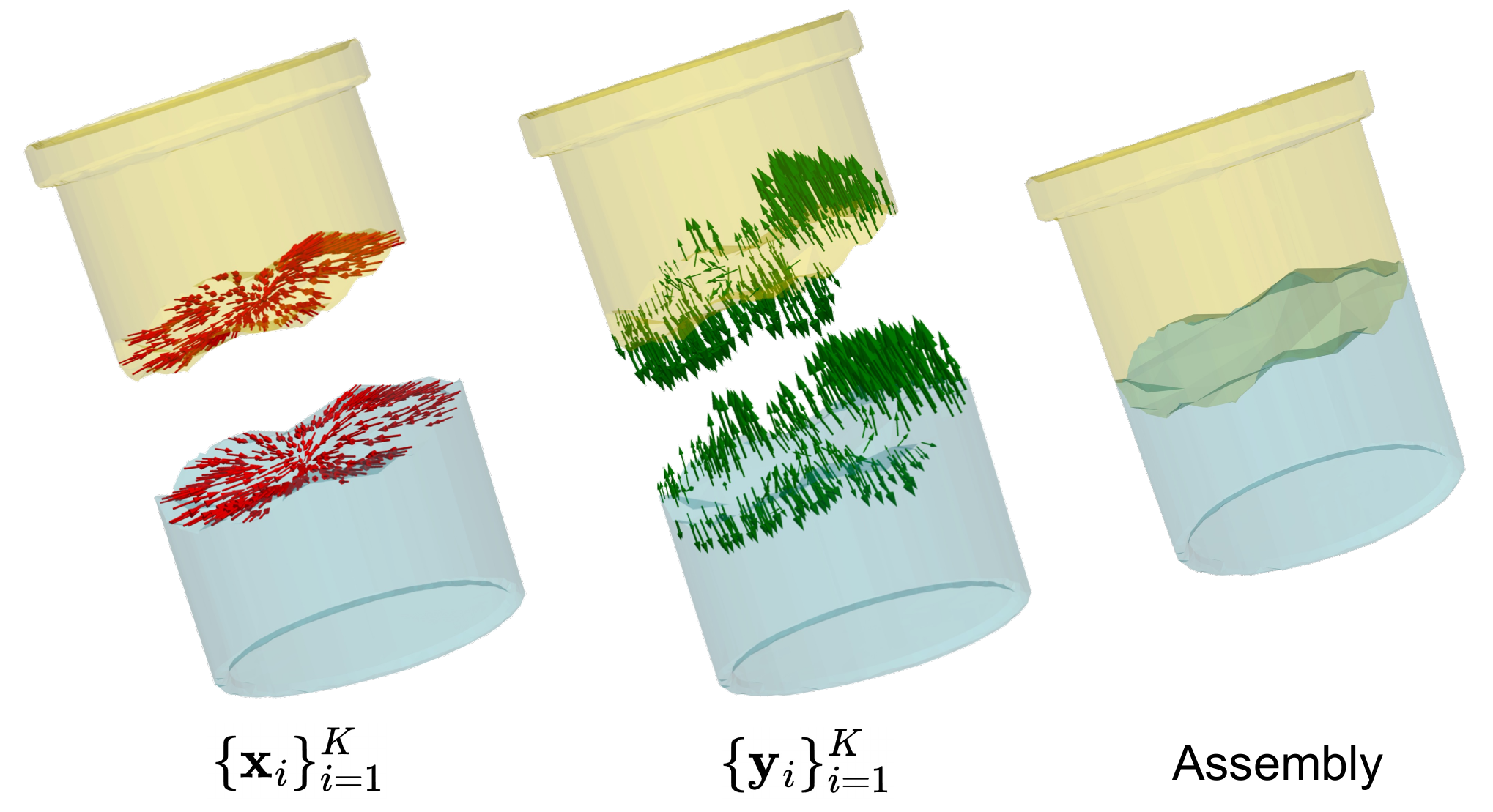}

\caption{\textbf{Visualization of learned orientations.} We visualize learned vectors of $\{\mathbf{x}_i\}_{i \in \mathcal{I}}$ (left, \textcolor[HTML]{CC0000}{red arrows}) and $\{\mathbf{y}_i\}_{i \in \mathcal{I}}$ (middle, \textcolor[HTML]{22741C}{green arrows}). The assembly result is shown on the right.}
\label{fig:orientation}
\vspace{-3mm}
\end{center}
\end{figure}

%% file: figures/heatmap.tex

\begin{figure}[t]
\begin{center}
\includegraphics[width=\linewidth]{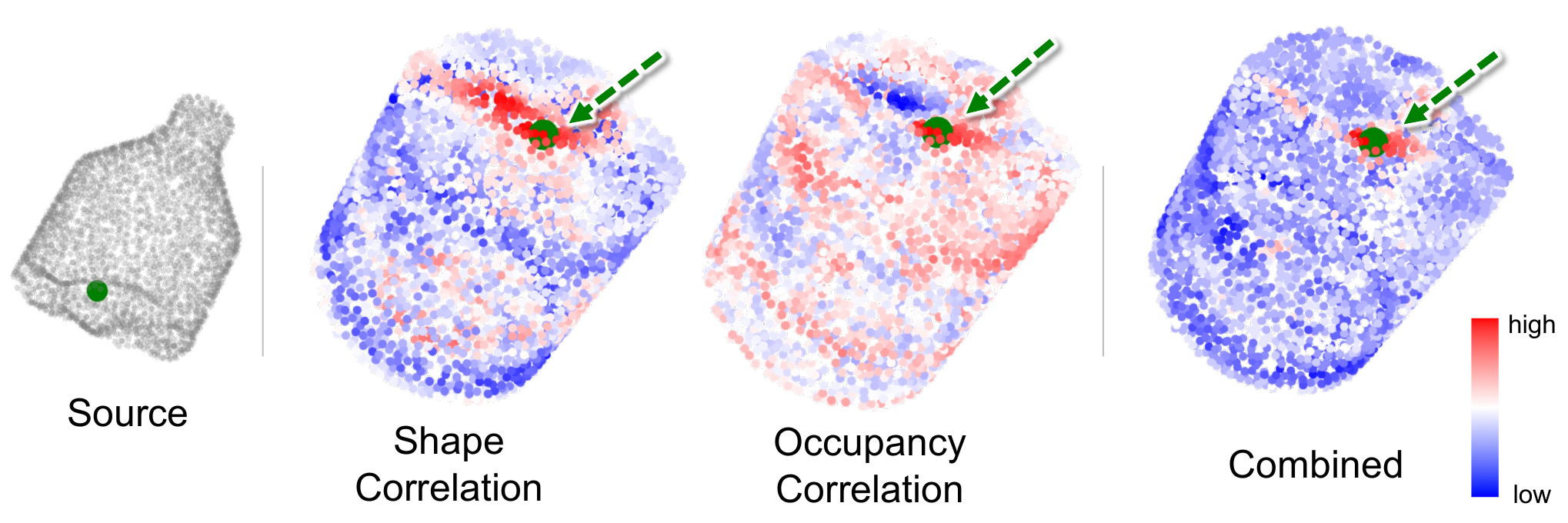}
\vspace{-3mm}
\caption{\textbf{Visualization of correlation distribution.} 
A green dot (\textcolor[HTML]{22741C}{$\bullet$}) on the left point cloud marks the source's $i$-th point, with corresponding true match points marked with green dots and arrows.
Point colors represent correlation score magnitudes for the $i$-th point's similarity to each target point, with \textcolor{red}{red} and \textcolor{blue}{blue} indicating high and low correlation scores, respectively.}
\label{fig:heatmap}
\vspace{-4mm}
\end{center}
\end{figure}

%% file: figures/feat_tsne.tex

\begin{figure*}[t]
\begin{center}
\includegraphics[width=\linewidth]{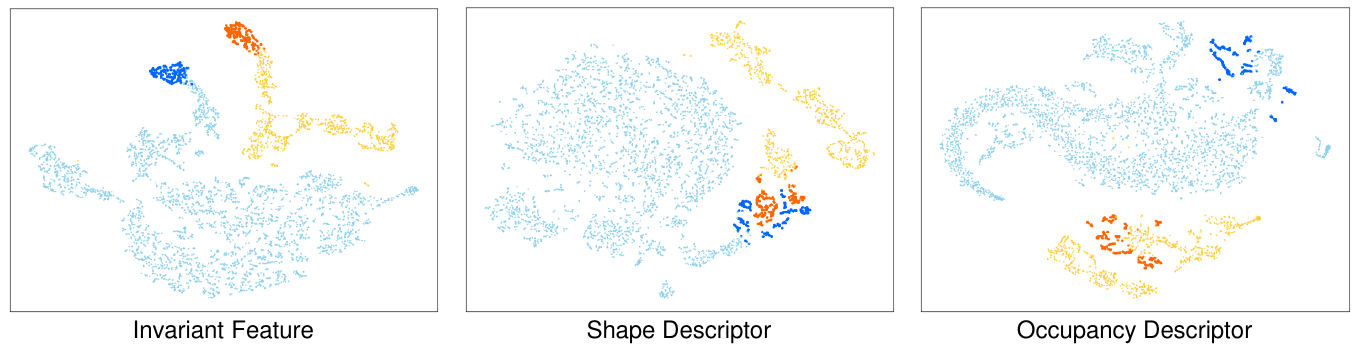}
\vspace{-4mm}
\caption{\textbf{Feature visualization via t-SNE.}
Mating surface points are displayed in blue (\textcolor{blue}{$\bullet$}) for the source and orange (\textcolor{orange}{$\bullet$}) for the target, while non-mating surface points are colored in skyblue (\textcolor{cyan}{$\bullet$}) for the source and yellow (\textcolor{yellow}{$\bullet$}) for the target.
}
\label{fig:tsne}
\end{center}
\vspace{-2mm}
\end{figure*}

%% file: tables/3_abl_combined.tex
\begin{table}[t]
    \vspace{-2mm}
    \centering
    \resizebox{\linewidth}{!}{
    \begin{tabular}{cc|rrrr}
        \toprule
        Occupancy & Orientation & CRD $\downarrow$ & CD $\downarrow$ & RMSE(R) $\downarrow$ & RMSE(T) $\downarrow$ \\
        Affinity & Loss ($\mathcal{L}_{\text{d}}$) & ($10^{-2}$) & ($10^{-3}$) & ($^{\circ}$) & ($10^{-2}$) \\
        \midrule 
        L2 dist & \textcolor{red}{\ding{55}} & 0.42 & 0.31 & 14.88 & 4.31 \\
        cosine & \textcolor{red}{\ding{55}} & 0.31 & 0.21 & 14.58 & 4.44 \\
        L2 dist & \textcolor[HTML]{22741C}{\ding{51}} & 0.38 & 0.30 & 13.29 & 3.81 \\
        cosine & \textcolor[HTML]{22741C}{\ding{51}} & \textbf{0.28} & \textbf{0.17} & \textbf{12.88} & \textbf{3.78} \\
        \bottomrule
    \end{tabular}
    }
    
    \makebox[\linewidth]{\footnotesize(a) Ablation study on combinative matching.} 

    \vspace{0.5em}
    \resizebox{\linewidth}{!}{
    \begin{tabular}{ccc|rrrr}
        \toprule
        Equivariant & Shape & Occupancy  & CRD $\downarrow$ & CD $\downarrow$ & RMSE(R) $\downarrow$ & RMSE(T) $\downarrow$ \\
        Embedding & Matching & Matching & ($10^{-2}$) & ($10^{-3}$) & ($^{\circ}$) & ($10^{-2}$) \\
        
        \midrule 
        \textcolor{red}{\ding{55}} & \textcolor[HTML]{22741C}{\ding{51}} & \textcolor[HTML]{22741C}{\ding{51}}  & 0.74 & 0.53 & 38.74 & 11.88 \\
        \textcolor[HTML]{22741C}{\ding{51}} & \textcolor{red}{\ding{55}} & \textcolor[HTML]{22741C}{\ding{51}}  & 0.38 & 0.28 & 13.17 & 3.86 \\
        \textcolor[HTML]{22741C}{\ding{51}} & \textcolor[HTML]{22741C}{\ding{51}}  & \textcolor{red}{\ding{55}} & 0.35 & 0.25 & 14.01 & 4.24 \\
        \textcolor[HTML]{22741C}{\ding{51}} & \textcolor[HTML]{22741C}{\ding{51}} & \textcolor[HTML]{22741C}{\ding{51}} & \textbf{0.28} & \textbf{0.17} & \textbf{12.88} & \textbf{3.78} \\

        \bottomrule
    \end{tabular}
    }
    \makebox[\linewidth]{\footnotesize (b) Ablation study on model components.} 
    \vspace{-5mm}
    \caption{Ablation studies of the proposed approach.}
    \vspace{-4mm}
    \label{table:res_real_data_wide}
\end{table}

%% file: tables/suppl_2_generalization.tex

\begin{table}[t]
    \centering
    \vspace{-2mm}
    \resizebox{\linewidth}{!}{
	\begin{tabular}{l|rrrr}
        \toprule
        \multirow{2}{*}{Method} & CRD $\downarrow$ & CD $\downarrow$ & RMSE(R) $\downarrow$ & RMSE(T) $\downarrow$ \\
        & ($10^{-2}$) & ($10^{-3}$) & ($^{\circ}$) & ($10^{-2}$)\\

        \cmidrule{1-5}
        \multicolumn{5}{c}{\texttt{everyday} $\rightarrow$ \texttt{artifact}}  \\
        \midrule
        Global~\cite{schor2019componet, li2020global} %
            & 19.68 & 6.37 & 85.08 & 21.48  \\
        LSTM~\cite{wu2020lstm} %
             & 19.45 & 6.37 & 84.07 & 21.24  \\
        DGL~\cite{huang2020dgl} %
             & 20.36 & 5.49 & 87.37 & 22.29 \\
        NSM~\cite{chen2022neural} %
            & 19.95 & 6.88 & 84.16 & 21.74 \\
        \citet{wu2023leveraging}  %
             & 19.13 & 7.98 & 85.27 & 22.96 \\
        GeoTransformer~\cite{qin2022geotransformer} %
            &  1.01 & 0.78 & 33.14 & 9.75\\
        Jigsaw~\cite{lu2023jigsaw}  %
            & 10.36 & 2.48 & 56.98 & 10.36 \\
        PMTR~\cite{lee2024pmtr} %
             & \underline{0.82} & \underline{0.59} & \underline{29.63} & \underline{9.21} \\
        \textbf{\oursabb~(Ours)} %
            & \textbf{0.74} & \textbf{0.54} & \textbf{25.67} & \textbf{7.73} \\ 

        \midrule
        \multicolumn{5}{c}{\texttt{artifact} $\rightarrow$ \texttt{everyday}}  \\
        \midrule
        Global~\cite{schor2019componet, li2020global} %
            & 20.78 & 10.23 & 86.74 & 23.06\\
        LSTM~\cite{wu2020lstm} %
             & 21.41 & 7.97 & 84.51 & 23.78\\
        DGL~\cite{huang2020dgl} %
             & 21.80 & 7.08 & 86.16 & 24.24\\
        NSM~\cite{chen2022neural} %
            & 21.34 & 8.52 & 85.46 & 23.58\\
        \citet{wu2023leveraging}  %
             & 20.70 & 11.67 & 85.81 & 22.96 \\
        GeoTransformer~\cite{qin2022geotransformer} %
            & 0.80 & 0.53 & 41.65 & 13.23\\
        Jigsaw~\cite{lu2023jigsaw}  %
            & 11.00 & 3.04 & 70.88 & 10.75\\
        PMTR~\cite{lee2024pmtr} %
            & \underline{0.64} & \textbf{0.44} & \underline{33.23} & \underline{10.97} \\
        \textbf{\oursabb~(Ours)} %
            & \textbf{0.62} & \underline{0.46} & \textbf{26.91} & \textbf{8.30} \\ 
        \bottomrule
        
	\end{tabular}
        \vspace{-8mm}
     }
    \caption{Transferability experiments on Breaking Bad~\cite{sellan2022breakingbad}.}
    \label{table:suppl_generalization}
    \vspace{-5mm}
\end{table}

%% file: tables/1_pairwise_bbad.tex

\begin{table}[t]
    \centering
    \resizebox{\linewidth}{!}{
	\begin{tabular}{l|rrrr}
        \toprule
        \multirow{2}{*}{Method} & CRD $\downarrow$ & CD $\downarrow$ & RMSE(R) $\downarrow$ & RMSE(T) $\downarrow$ \\
        & ($10^{-2}$) & ($10^{-3}$) & ($^{\circ}$) & ($10^{-2}$)\\

        \cmidrule{1-5}
        \multicolumn{5}{c}{\texttt{everyday}}  \\
        \midrule
        Global~\cite{schor2019componet, li2020global} %
            & 27.77 & 15.26 & 110.74 & 30.61  \\
        LSTM~\cite{wu2020lstm} %
             & 20.04 & 7.77 & 84.60 & 22.07  \\
        DGL~\cite{huang2020dgl} %
             & 20.32 & 6.40 & 86.23 & 22.38  \\
        NSM~\cite{chen2022neural} %
            & 21.71 & 11.09 & 83.38 & 23.71 \\
        \citet{wu2023leveraging}  %
             & 20.65 & 11.66 & 84.58 & 22.90  \\
        GeoTransformer~\cite{qin2022geotransformer} %
            & 0.61 & 0.51 & 22.81 & 7.28 \\
        Jigsaw~\cite{lu2023jigsaw}  %
            & 5.48 & 1.34 & 38.73 & \textbf{2.73}  \\
        PMTR~\cite{lee2024pmtr} %
             & \underline{0.39} & \underline{0.25} & \underline{17.14} & 5.53 \\
        \textbf{\oursabb~(Ours)} %
            & \textbf{0.28} & \textbf{0.17} & \textbf{12.88} & \underline{3.78} \\ 

        \midrule
        \multicolumn{5}{c}{\texttt{artifact}}  \\
        \midrule
        Global~\cite{schor2019componet, li2020global} %
            & 19.26 & 7.16 & 86.30 & 21.02  \\
        LSTM~\cite{wu2020lstm} %
             & 19.52 & 6.45 & 84.42 & 21.33  \\
        DGL~\cite{huang2020dgl} %
             & 19.82 & 6.19 & 85.46 & 21.65 \\
        NSM~\cite{chen2022neural} %
            & 19.44 & 6.33 & 83.22 & 21.41 \\
        \citet{wu2023leveraging}  %
             & 19.17 & 7.97 & 85.04 & 20.90  \\
        GeoTransformer~\cite{qin2022geotransformer} %
            & 0.89 & 0.70 & 33.23 & 10.30 \\
        Jigsaw~\cite{lu2023jigsaw}  %
            & 6.36 & 1.45 & 39.71 & \textbf{3.02}  \\
        PMTR~\cite{lee2024pmtr} %
             & \underline{0.60} & \underline{0.42} & \underline{23.28} & 7.27 \\
        \textbf{\oursabb~(Ours)} %
            & \textbf{0.49} & \textbf{0.34} & \textbf{18.77} & \underline{5.57} \\ 
        \bottomrule
        
	\end{tabular}
        \vspace{-7mm}
     }
    \caption{Pairwise shape assembly results. Numbers in \textbf{bold} indicate the best performance and \underline{underlined} ones are the second best.}
    \label{table:pairwise_v3}
    \vspace{-3.0mm}
\end{table}

%% file: figures/pairwise_qual.tex

\begin{figure}[t]
\begin{center}
\includegraphics[width=\linewidth]{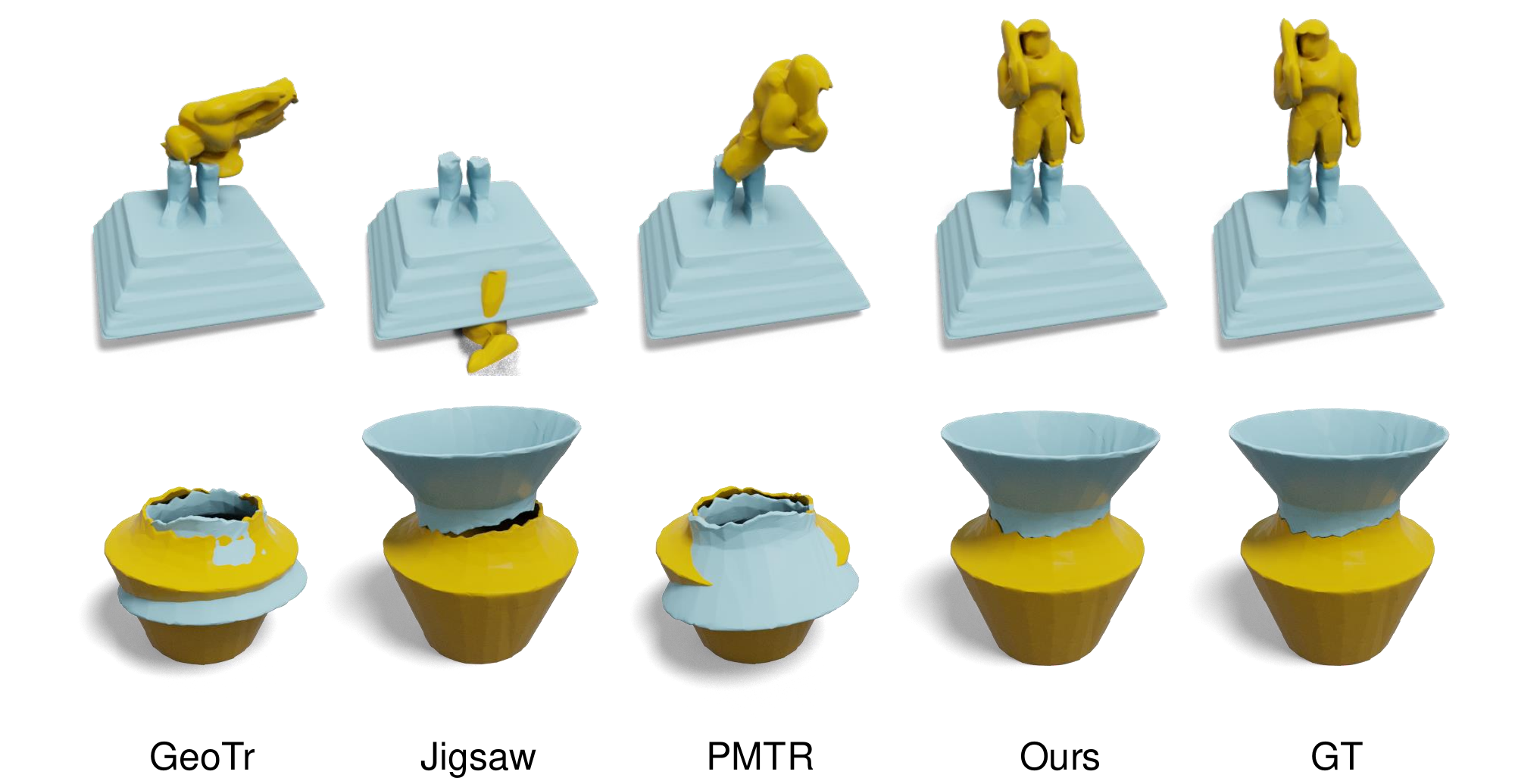}
\vspace{-5mm}
\caption{Qualitative comparison for pairwise shape assembly.}
\label{fig:pairwise_qual}
\vspace{-7mm}
\end{center}
\end{figure}

%% file: tables/2_mpa_bbad.tex

\begin{table}[t]
    \centering
    \resizebox{\linewidth}{!}{
	\begin{tabular}{l|rrrrrr}
        \toprule
        \multirow{2}{*}{Method} & CRD $\downarrow$ & CD $\downarrow$ & RMSE(R) $\downarrow$ & RMSE(T) $\downarrow$ & PA$_{\text{CRD}}$ $\uparrow$ & PA$_{\text{CD}}$ $\uparrow$\\
        & ($10^{-2}$) & ($10^{-3}$) & ($^{\circ}$) & ($10^{-2}$) & (\%) & (\%)\\

        \cmidrule{1-7}
        \multicolumn{7}{c}{\texttt{everyday}}  \\
        \midrule
        Global~\cite{schor2019componet, li2020global} %
            & 27.79 & 15.30 & 55.42 & 15.31 & 36.42 & 37.90 \\
        LSTM~\cite{wu2020lstm} %
             & 27.69 & 15.23 & 54.78 & 15.24 & 36.74 & 38.97  \\
        DGL~\cite{huang2020dgl} %
             & 27.90 & 13.23 & 55.76 & 15.33 & 36.99 & 39.70 \\
        \citet{wu2023leveraging}  %
             & 28.18 & 19.70 & 54.98 & 15.59 & 35.66 & 36.28  \\
        Jigsaw~\cite{lu2023jigsaw}  %
            & 14.13 & 11.82 & 41.12 & 11.74 & 52.48 & 60.26 \\
        PMTR~\cite{lee2024pmtr} %
             & \underline{6.51} & \underline{5.56} & \underline{31.57} & \underline{9.95} & \underline{66.95} & \underline{70.56} \\
        \textbf{\oursabb~(Ours)} %
            & \textbf{5.18} & \textbf{3.65} & \textbf{27.11} & \textbf{8.13} & \textbf{73.88} & \textbf{77.88} \\

        \midrule
        \multicolumn{7}{c}{\texttt{artifact}}  \\
        \midrule
        Global~\cite{schor2019componet, li2020global} %
            & 26.42 & 14.92 & 54.41 & 14.48 & 36.67 & 36.97 \\
        LSTM~\cite{wu2020lstm} %
             & 28.15 & 14.61 & 53.59 & 15.49 & 36.67 & 37.25 \\
        DGL~\cite{huang2020dgl} %
             & 27.48 & 13.91 & 54.66 & 15.10 & 36.66 & 37.40 \\
        \citet{wu2023leveraging}  %
             & 26.02 & 15.81 & 54.35 & 14.27 & 36.63 & 37.02 \\
        Jigsaw~\cite{lu2023jigsaw}  %
            & 16.10 & 9.53 & 42.01 & 17.47 & 56.93 & 65.58  \\
        PMTR~\cite{lee2024pmtr} %
             & \underline{5.67} & \underline{4.33} & \underline{31.58} & \underline{10.08} & \underline{66.96} & \underline{71.61} \\
        \textbf{\oursabb~(Ours)} %
            & \textbf{4.56} & \textbf{3.04} & \textbf{29.21} & \textbf{8.99} & \textbf{71.02} & \textbf{76.32} \\ 
        \bottomrule
        
	\end{tabular}
    }
    \caption{Multi-part assembly results. Numbers in \textbf{bold} indicate the best performance and \underline{underlined} ones are the second best.}
    \vspace{-2mm}
    \label{table:mpa}
\end{table}

%% file: figures/mpa_qual.tex

\begin{figure}[t]
\begin{center}
\includegraphics[width=\linewidth]{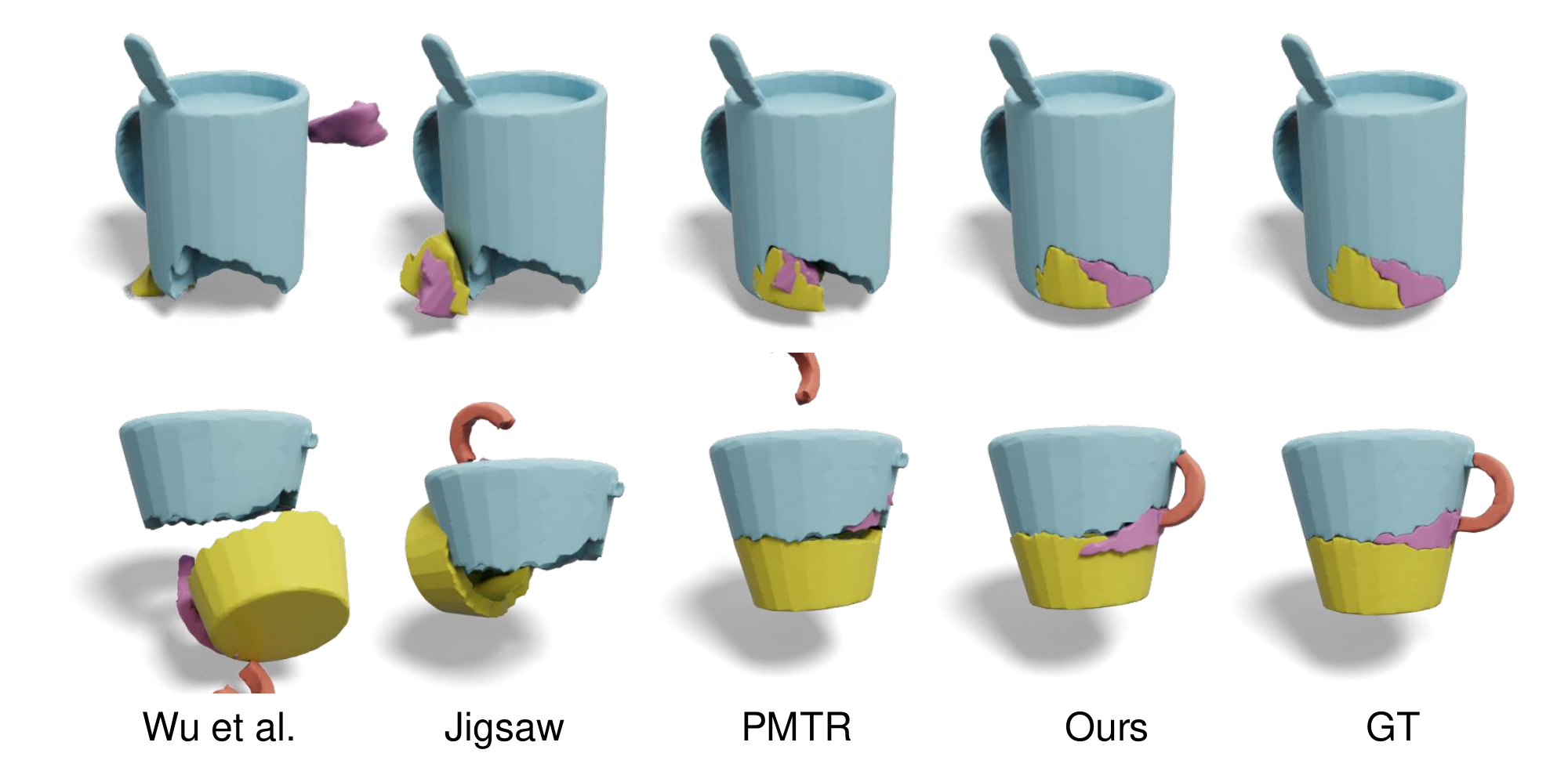}
\vspace{-6mm}
\caption{Qualitative comparison for multi-part assembly.}
\label{fig:mpa_qual}
\end{center}
\vspace{-6mm}
\end{figure}

%% file: sections/5_conclusion.tex

\section{Limitations and Future Work}
\label{sec:limitation}
While our method demonstrates robust performance in most assembly scenarios, it still fails in certain challenging scenarios, as illustrated in Fig.~\ref{fig:rebuttal_failure_case_and_multi_identical}.
First, given extremely low overlap between mating surfaces (a,b), the occupancy cues become too weak to provide reliable guidance.
Second, when fracture surfaces are visually indistinguishable (b,c), the pairwise matching scores become less discriminative, often resulting in incorrect part permutations.
Integrating additional information such as texture \& color cues, or enforcing cycle-consistency, could address such ambiguities.
\input{figures/failure_case}

\section{Conclusion}
\label{sec:conclusion}
We have introduced combinative matching that incorporates the multi-faceted, task-oriented properties, which demonstrated the superiority over recent baselines by capturing intrinsic properties of assembly, such as degrees of convexity/concavity and orientation of mating surfaces, even without explicit supervision.
Although this paper explores learning orientation, shape, and occupancy matching, the method can be further expanded to incorporate properties like physical compatibility or functional constraints, paving the way for more versatile assembly frameworks.

%% file: figures/failure_case.tex

\begin{figure}[h]
  \centering
        \centering
        \includegraphics[width=0.95\linewidth]{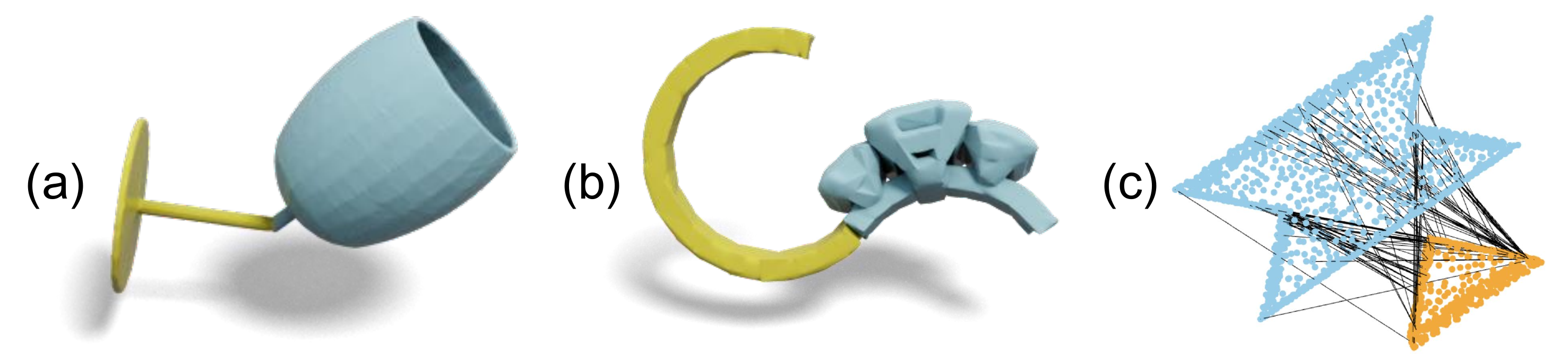}
        \vspace{-3mm}
      \caption{\textbf{(a-b)} Representative failure cases on Breaking Bad. \textbf{(c)} Visualization of top-$k$ matches on a toy example ($k=128$).}
      \label{fig:rebuttal_failure_case_and_multi_identical}
      \vspace{-4mm}
\end{figure}

%% file: sections/6_acknowledgement.tex
\section*{Acknowledgements}

This work was supported by IITP grants (RS-2022-II220290: Visual Intelligence for Space-Time Understanding \& Generation (30\%), RS-2021-II212068: AI Innovation Hub (35\%), RS-2024-00457882: National AI Research Lab Project (30\%), RS-2019-II191906: AI Graduate School Program at POSTECH (5\%)) funded by Ministry of Science and ICT, Korea.
Additionally, we would like to express our gratitude to Samsung Research America (SRA) NA-AIC for their invaluable support.

%% file: sections/X_suppl.tex

\clearpage

\renewcommand{\thesection}{\Alph{section}}
\appendix
\setcounter{table}{0}
\setcounter{figure}{0}
\maketitlesupplementary

\input{sections/suppl/0_intro}
\input{sections/suppl/1_exp_result}
\input{sections/suppl/2_model_details}
\input{sections/suppl/3_exp_setup}

\input{figures/suppl_pairwise_qual}

\input{figures/suppl_mpa_qual}

%% file: sections/suppl/0_intro.tex
\renewcommand{\thefigure}{A\arabic{figure}}
\renewcommand{\thetable}{A\arabic{table}}

\noindent
In this supplementary material, we present additional information and analyses not included in the main paper.
In Section~\ref{sec:additional_exp}, we provide further analysis of Combinative Matching using a toy dataset as well as extended experimental results and analyses, including further analysis on learned descriptors and orientations, as well as additional experimental results.
In Section~\ref{sec:additional_method_detail}, we detail the network architecture, including equivariant feature extractor, orientation hypothesizer, invariant feature computation, matching modules, and training objectives.
In Section~\ref{sec:additional_setup}, we describe additional details on the training and evaluation recipes, including hyperparameter settings and the evaluation details for multi-part assembly.

%% file: sections/suppl/1_exp_result.tex

\section{Additional Experimental Results}
\label{sec:additional_exp}

\subsection{Results on Vanilla Breaking Bad Dataset}

Since all our experiments were conducted on the volume-constrained version of the Breaking Bad dataset~\cite{sellan2022breakingbad}, we additionally provide a quantitative comparison on the vanilla version of the Breaking Bad dataset for a fair comparison with prior methods.
\input{tables/suppl_8_vanilla}

Table~\ref{table:vanilla_bbad} shows that ours consistently achieves accurate assembly, outperforming previous state-of-the-art methods. 
The robustness of our approach is particularly evident in cross-subset evaluation (\texttt{everyday} $\rightarrow$ \texttt{artifact}), where the performance remains stable despite substantial variations in object categories and fragment characteristics.

\subsection{Further Analysis on Combinative Matching}
\label{sec:suppl_ambiguous}

In this section, we provide deep, but intuitive analyses of our combinative matching, highlighting the necessity of explicit occupancy learning for robust shape assembly.
To illustrate its significance, we build a synthetic dataset that highlights the critical role of occupancy learning, particularly in scenarios with ambiguous geometric patterns, \eg, visual resemblance shown in Fig.~\ref{fig:ambiguous}.

\subsubsection{Toy dataset with local ambiguity}
\label{sec:toy_dataset}
\input{figures/suppl_ambiguous}

\input{figures/suppl_pattern}

\noindent\textbf{Ambiguity patterns.}
The synthetic dataset consists of six carefully designed patterns, as shown in Fig.~\ref{fig:pattern}. 
These patterns were deliberately crafted to ensure that correct assembly depends primarily on recognizing occupancy and directionality rather than visual resemblance.
We use pattern 1\verb|~|3 for training, enabling the model to learn occupancy relationships and effectively mitigate visual ambiguities explicitly.
Then patterns 4\verb|~|6 are used to validate our assumption: if the model successfully learns occupancy-based complementary relationships from the train set, it should inherently generalize well to the visually analogous but structurally different test patterns.

We generate 200 random objects of each pattern \{1, 2, 3\} for training, 50 objects of each for validation, and 50 objects of each pattern \{4, 5, 6\} for testing.
Although each pattern retains its pattern structure, small shape variations are introduced randomly.

\subsubsection{Experiments}

\input{tables/suppl_4_toy_result}
\input{tables/suppl_5_toy_abl_ours}

Table~\ref{table:toy_quan} compares our method against recent state-of-the-art methods, \eg, Jigsaw~\cite{lu2023jigsaw} and PMTR~\cite{lee2024pmtr}, on the synthetic dataset.
In particular, both Jigsaw~\cite{lu2023jigsaw} and PMTR~\cite{lee2024pmtr} show limited performance when tackling parts that exhibit visually similar but occupancy-opposed surfaces.
By contrast, our method enforces a direct contrast in volume occupancy alongside shape similarity, thereby achieving robust, interlocking matches and significantly improving alignment accuracy across all metrics.

\smallbreak\noindent\textbf{Ablation Study.}
In addition to comparing against existing methods, we validate the effect of explicit occupancy matching by ablating our framework.
Table~\ref{table:ambiguous_cm} reports the performance when removing the occupancy branch and relying solely on shape-based similarity.
This confirms that identifying visually similar surfaces alone is not sufficient:  
\emph{opposite} volume occupancy must also be enforced to prevent misleading matches arising from superficially alike shapes.

\begin{figure}[ht]
\begin{center}
\includegraphics[width=\linewidth]{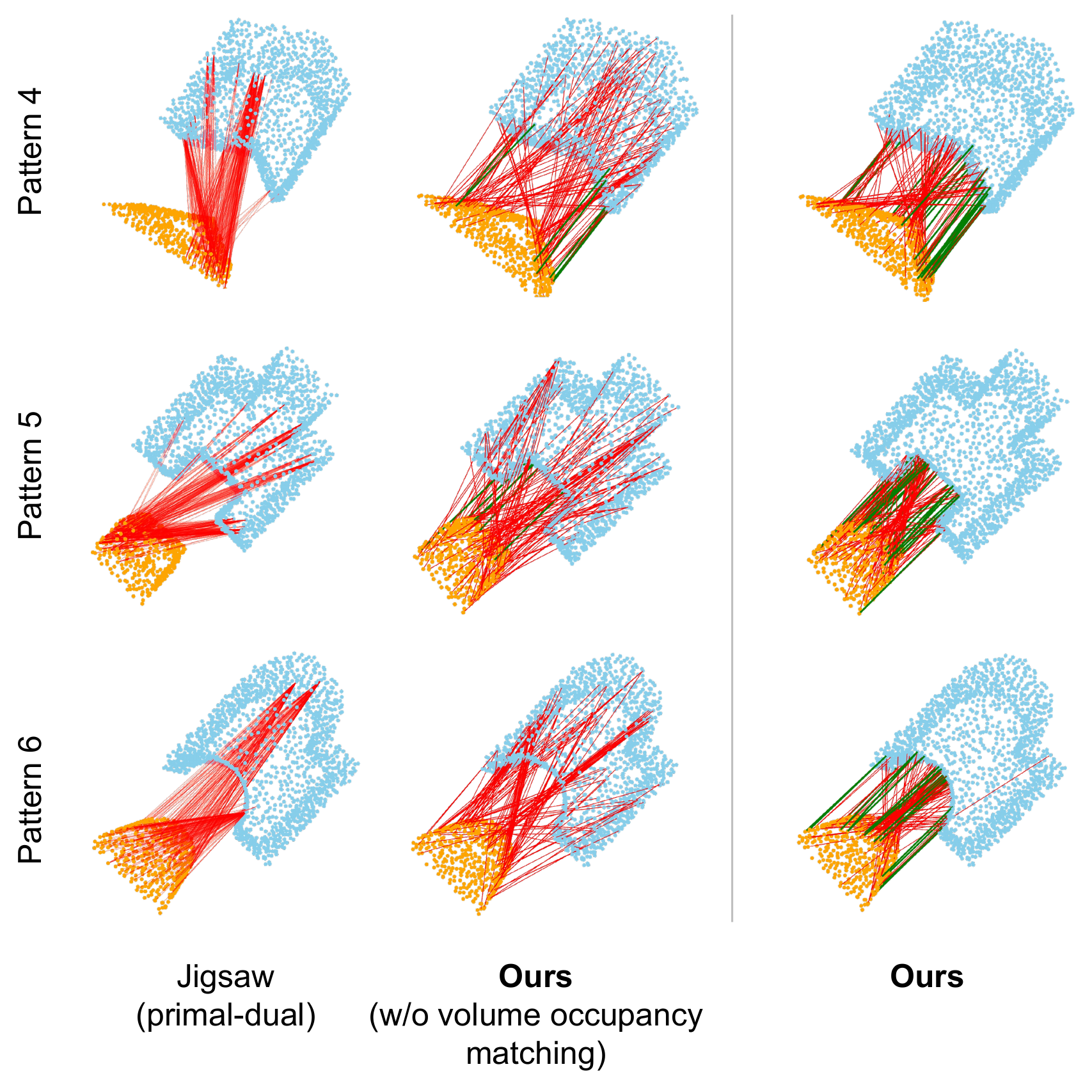}
\vspace{-3mm}
\caption{Visualization of top-$k$ matches ($k=128$). 
Positive matches are colored in green, while negative matches are colored in red.}
\label{fig:suppl_match}
\end{center}
\vspace{-7mm}
\end{figure}

\input{tables/suppl_6_toy_abl_jigsaw}

\noindent\textbf{Discussion on Jigsaw~\cite{lu2023jigsaw}.}
Jigsaw~\cite{lu2023jigsaw} applies a primal-dual descriptor specifically to the predicted mating surfaces, aiming to distinguish convex regions from concave ones.  
However, Table~\ref{table:ambiguous_jigsaw} shows that whether Jigsaw uses single-descriptor or primal-dual streams, and regardless of whether it segments mating surfaces or not, the improvements over a naive approach remain marginal.  
In several cases, primal-dual matching even underperforms single matching, indicating that the method struggles to capture a genuine complementary relationship (\ie, opposing occupancy).  
We additionally provide ground-truth surface masks (GT) as a further upper bound, yet still see little gain from dual learning.  
Visualization result of top-$k$ matches in Fig.~\ref{fig:suppl_match} corroborate this: the two-stream descriptor alone does not robustly reject visually deceptive matches, confirming that an explicit “identical shape + opposite occupancy” objectives are key to resolving ambiguous interfaces.

\input{tables/suppl_7_toy_abl_pmtr}

\smallbreak
\noindent\textbf{Discussion on PMTR~\cite{lee2024pmtr}.}
As shown in Tab.~\ref{table:ambiguous_pmtr}, whether PMTR employs only a coarse matcher (PMT) or includes an additional fine matcher, both settings still struggle to address patterns demanding explicit occupancy opposition.
While fine matching often helps localize small-scale interfaces in standard registration tasks, here it does not substantially alleviate the core limitation of PMTR’s design: maximizing only \textit{identical surface shape} similarity.
Without negatively enforcing “occupancy,” the method remains prone to errors on visually alike yet geometrically mismatched parts.

\input{figures/suppl_heatmap}

\subsection{Additional orientation analysis}
As discussed in our main paper (Fig.~\color{red}4\color{black}~in Sec.~\color{red}4.3\color{black}), the learned orientations, $\mathbf{F}_{\text{d}}^{\text{P}}$ and $\mathbf{F}_{\text{d}}^{\text{Q}}$, exhibit several notable patterns that enable the model to effectively align mating surface along with high interpretability.
To further analyze these learned patterns across various examples, we provide additional visualizations of learned orientation under the same experimental setups described in Sec.~\color{red}4.3\color{black}.

Figure~\ref{fig:suppl_orientation} presents the results for six distinct objects, again showcasing consistent patterns:
(1) parallel alignment of source and target orientations ($\mathbf{x}_i$ and $\mathbf{y}_i$), (2) $\mathbf{x}_i$ directed toward the center of mating surfaces, (3) parallel alignment of $\mathbf{x}_i$ with 2D plane of mating surfaces, (4) outward/inward directed $\mathbf{y}_i$ based on convexity/concavity, and (5) correlation between magnitudes of $\mathbf{y}_i$ and surface curvature.
These patterns highlight the robustness and adaptability of our method in learning {\em valid orientations} that respect the geometry and complementarity of mating surfaces {\em without any explicit supervision}, verifying both the effectiveness and interpretability of the proposed combinative matching.

\subsection{Additional correlation heatmap analysis}
As in Fig.~\color{red}5\color{black}~of Sec.~\color{red}4.3\color{black}, we compare the correlation matrices for shape $\mathbf{C}_{\text{s}}$, occupancy $\mathbf{C}_{\text{o}}$, and combined correlation $\mathbf{C}$ of additional examples to further validate the efficacy of the proposed combinative matching approach over the conventional matching.
Following the same experimental setup described Sec.~\color{red}4.3\color{black}~of the main paper, Fig.~\ref{fig:suppl_heatmap} illustrates the comparison, where we observe similar phenomena to those presented in the main paper.

When relying solely on the shape distribution $(\mathbf{C}_{\text{s}})_i$, the best target match for the $i$-th source point is distributed over multiple regions due to local ambiguity of visual resemblance.
The occupancy distribution $(\mathbf{C}_{\text{o}})_i$ reveals relatively uniform scores across the surface, with a slightly higher concentration near the true match, offering complementary information but lacking precise localization.
Integrating shape and occupancy information effectively resolves both local ambiguity and match confidence uncertainty, highlighting the importance of task-oriented multiple representation learning in combinative matching.

\input{figures/suppl_orientation}

\subsection{Additional qualitative results}
We provide additional qualitative comparisons against recent state-of-the-art methods~\cite{qin2022geotransformer,wu2023leveraging,lu2023jigsaw,lee2024pmtr} on the Breaking Bad dataset.
Figures~\ref{fig:suppl_pairwise_qual} and~\ref{fig:suppl_mpa_qual} illustrate the comparison results in pairwise and multi-part settings, respectively.
From the baselines, several notable patterns emerge:
(1) \textbf{Failure in localizing mating surfaces}:
The baselines lack an understanding of local orientations and occupancy on the mating surfaces.
This results in incorrect placement of parts, which are often located in the air rather than at the interfaces of their corresponding parts.
Examples of this failure include (a,c,d,e,f,h,i,j,k,n)-Wu~\etal~\cite{wu2023leveraging}, (c,d,f,g,h,i,k)-Jigsaw~\cite{lu2023jigsaw}, and (d,g)-PMTR~\cite{lee2024pmtr} as shown in Fig.~\ref{fig:suppl_mpa_qual}. \\
(2) \textbf{Failure in establishing correct correspondence}:
While some methods perform decent localization of mating surfaces in pairwise assembly, they often fail to establish accurate correspondences due to local ambiguities, thus leading to incorrect assembly configurations such as reversed or overlapping parts.
Specific examples of this issue are (d,g,h,k)-GeoTr~\cite{qin2022geotransformer}, (b,i,m)-Jigsaw~\cite{lu2023jigsaw}, and (a,c,e,f,h,j,k)-PMTR~\cite{lee2024pmtr}, as observed in Fig.~\ref{fig:suppl_pairwise_qual}.
These observations highlight critical challenges faced by existing methods in both accurate interface localization and resolving ambiguities during assembly.
By addressing these issues, we show that the proposed combinative matching demonstrates superior quantitative and qualitative results in both pairwise and multi-part scenarios.

%% file: tables/suppl_8_vanilla.tex

\begin{table}[H]
    \centering
    \resizebox{\linewidth}{!}{
	\begin{tabular}{l|cccc}
        \toprule
        \multirow{2}{*}{Method} & RMSE(R) $\downarrow$ & RMSE(T) $\downarrow$ & PA$_{\text{CD}}$ $\uparrow$ & CD $\downarrow$ \\
        & ($^{\circ}$) & ($10^{-2}$) & ($\%$)  &  ($10^{-3}$) \\
        
        \cmidrule{1-5}
        \multicolumn{5}{c}{\texttt{everyday}}  \\
        \midrule
        Global~\cite{schor2019componet, li2020global} %
            & 80.7 & 15.1 & 24.6 & 14.6 \\
        LSTM~\cite{wu2020lstm} %
             & 84.2 & 16.2 & 22.7 & 15.8 \\
        DGL~\cite{huang2020dgl} %
             & 79.4 & 15.0 & 31.0 & 14.3 \\
        \citet{wu2023leveraging}  %
             & 79.3 & 16.9 & 8.41 & 28.5 \\
        
        DiffAssemble~\cite{scarpellini2024diffassemble} %
            & 73.3 & 14.8 & 27.5 & - \\
        Jigsaw~\cite{lu2023jigsaw}  %
            & 42.3 & 10.7 & 57.3 & 13.3 \\
        PuzzleFusion++~\cite{wang2024puzzlefusion++} %
             & \underline{38.1} & \textbf{8.0} & \underline{71.0} & \underline{6.0} \\
        \textbf{\oursabb~(Ours)} %
            & \textbf{32.0} & \underline{9.6} & \textbf{77.3} & \textbf{3.5} \\ 

        \cmidrule{1-5}
        \multicolumn{5}{c}{\texttt{everyday $\rightarrow$ artifact}}  \\
        \cmidrule{1-5}
        Jigsaw~\cite{lu2023jigsaw}  %
            & 52.4 & 22.2 & 45.6 & \underline{14.3} \\
        PuzzleFusion++~\cite{wang2024puzzlefusion++} %
             & \underline{52.1} & \textbf{13.9} & \underline{49.6} & 14.5 \\
        \textbf{\oursabb~(Ours)} %
            & \textbf{46.0} & \underline{14.3} & \textbf{52.6} & \textbf{9.8} \\ 
            
        \bottomrule
	\end{tabular}
        \vspace{-7mm}
     }
    \caption{Multi-part assembly results on vanilla Breaking Bad dataset~\cite{sellan2022breakingbad}. Numbers in \textbf{bold} indicate the best performance and \underline{underlined} ones are the second best.}
    \label{table:vanilla_bbad}
\end{table}

%% file: figures/suppl_ambiguous.tex

\begin{figure}[!h]
\begin{center}
\includegraphics[width=\linewidth]{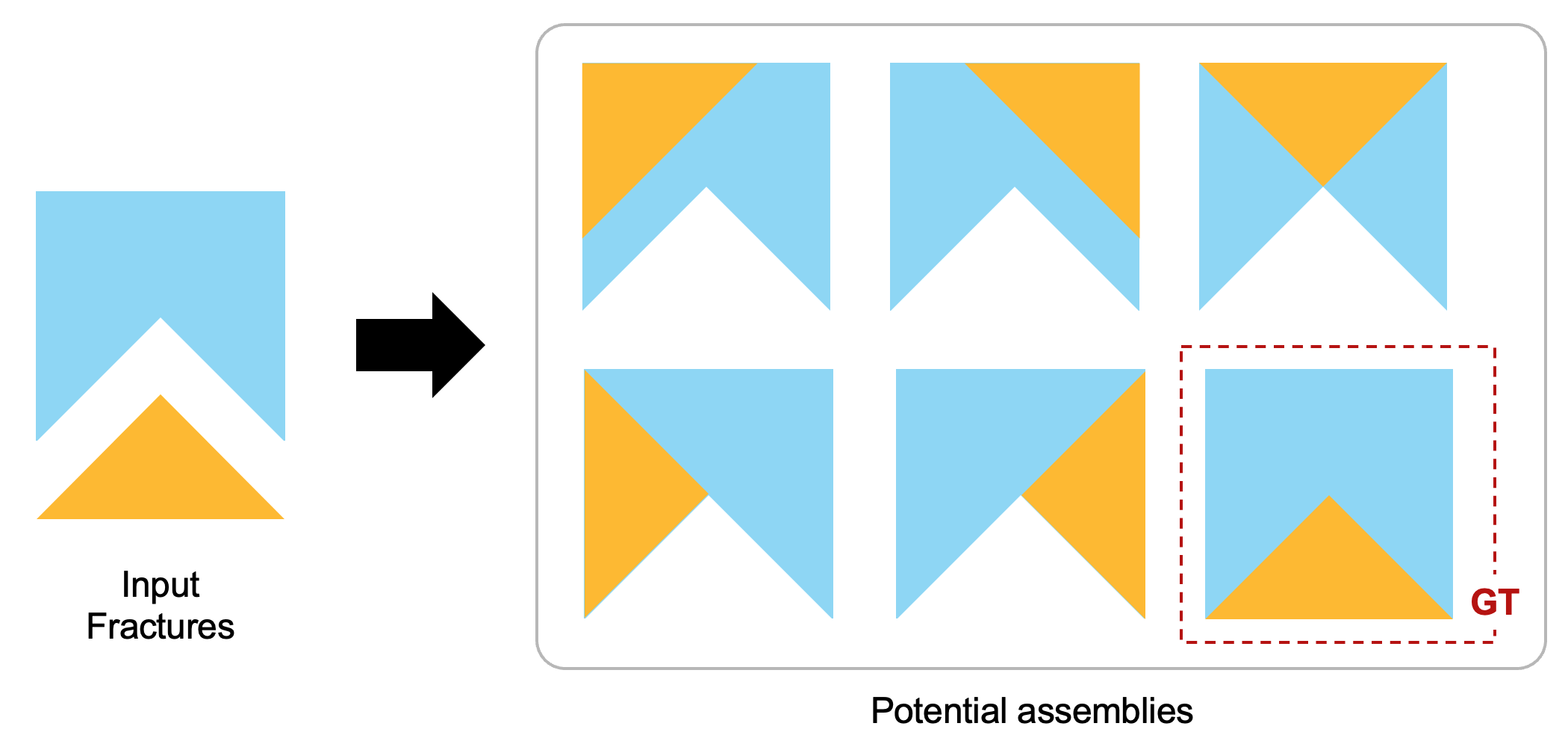}

\caption{Example of potential failure assemblies caused by \textbf{\textit{visual ambiguity}} within matching. }
\vspace{-5mm}
\label{fig:ambiguous}
\end{center}
\end{figure}

%% file: figures/suppl_pattern.tex

\begin{figure}[!h]
\begin{center}
\includegraphics[width=\linewidth]{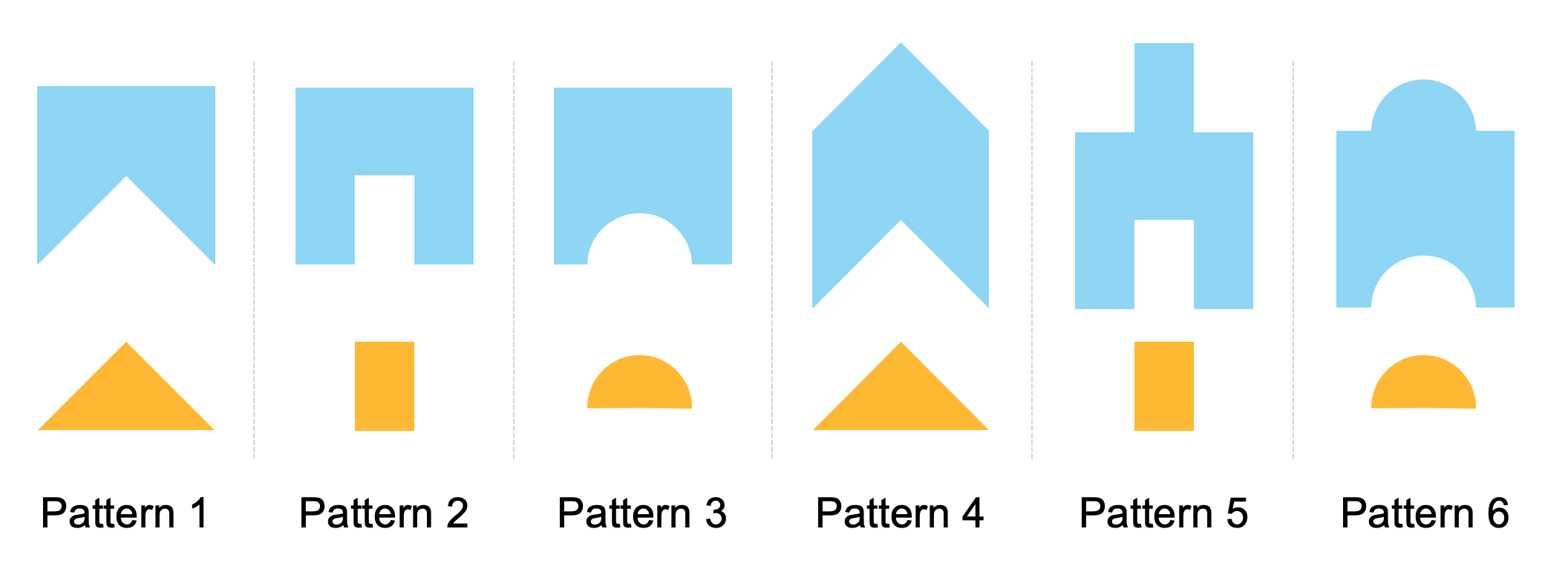}

\caption{Six types of ambiguity pattern for toy dataset. 2D polygons are further extruded to 3D meshes.}
\vspace{-5mm}
\label{fig:pattern}
\end{center}
\end{figure}

%% file: tables/suppl_4_toy_result.tex
\begin{table}[h]
    \centering
	\resizebox{.8\linewidth}{!}{
	\begin{tabular}{c|cccc}
        \toprule
         \multirow{2}{*}{Method} & CRD $\downarrow$ & CD $\downarrow$ & RMSE(R) $\downarrow$ & RMSE(T) $\downarrow$ \\
         & ($10^{-2}$) & ($10^{-3}$) & ($^{\circ}$) & ($10^{-2}$) \\
        \midrule 
        Jigsaw~\cite{lu2023jigsaw} & 17.64 & 6.95 & 84.98 & 24.81 \\
        PMTR~\cite{lee2024pmtr} & \underline{16.01} & \underline{6.01} & \underline{70.13} & \underline{15.70}  \\
        \textbf{\oursabb~(Ours)} & \textbf{9.24} & \textbf{2.07} & \textbf{55.29} & \textbf{13.40} \\
        \bottomrule
	\end{tabular}
	}
    \vspace{-2mm}
    \caption{
		Pairwise shape assembly results on toy dataset. 
        }
    \label{table:toy_quan}
\end{table}

%% file: tables/suppl_5_toy_abl_ours.tex
\begin{table}[h]
    \centering
	\resizebox{.99\linewidth}{!}{
	\begin{tabular}{cc|cccc}
        \toprule
         Shape & Occupancy  & CRD $\downarrow$ & CD $\downarrow$ & RMSE(R) $\downarrow$ & RMSE(T) $\downarrow$ \\
        Matching & Matching & ($10^{-2}$) & ($10^{-3}$) & ($^{\circ}$) & ($10^{-2}$) \\
        
        \midrule 
         {\ding{51}}  & & 11.95 & 3.92 & 66.90 & 17.06 \\
        {\ding{51}} & {\ding{51}} & \textbf{9.24} & \textbf{2.07} & \textbf{55.29} & \textbf{13.40} \\

        \bottomrule
	\end{tabular}
	}
    \vspace{-2mm}
    \caption{
		Ablation study on matching strategy.}
    \label{table:ambiguous_cm}
\end{table}

%% file: tables/suppl_6_toy_abl_jigsaw.tex
\begin{table}[h]
    \centering
    \resizebox{\linewidth}{!}{
    \begin{tabular}{cc|cccc}
        \toprule
        Surface & Desc. & CRD $\downarrow$ & CD $\downarrow$ & RMSE(R) $\downarrow$ & RMSE(T) $\downarrow$ \\
        Segmentation & Type & ($10^{-2}$) & ($10^{-3}$) & ($^{\circ}$) & ($10^{-2}$) \\
        \midrule
        & single & \textbf{17.24} & 7.48 & 87.84 & \textbf{21.84} \\
        & primal-dual & 17.80 & \textbf{6.74} & \textbf{87.47} & 23.17 \\
        \midrule
        \multirow{2}{*}{\ding{51}} & single & \textbf{17.19} & 7.29 & 89.06 & \textbf{21.68} \\
        & primal-dual & 17.64 & \textbf{6.95} & \textbf{84.98} & 24.81 \\
        \midrule
        \multirow{2}{*}{GT} & single & \textbf{12.88} & \textbf{4.63} & \textbf{71.10} & \textbf{10.89} \\
        & primal-dual & 13.04 & 5.63 & 72.04 & 11.98 \\
        \bottomrule
    \end{tabular}
    }
    \caption{
		Ablation study on surface segmentation and primal-dual matching modules in Jigsaw~\cite{lu2023jigsaw}.
        }
    \label{table:ambiguous_jigsaw}
\end{table}

%% file: tables/suppl_7_toy_abl_pmtr.tex
\begin{table}[h]
    \centering
	\resizebox{.99\linewidth}{!}{
	\begin{tabular}{cc|cccc}
        \toprule
        Coarse & Fine  & CRD $\downarrow$ & CD $\downarrow$ & RMSE(R) $\downarrow$ & RMSE(T) $\downarrow$ \\
        Matcher & Matcher & ($10^{-2}$) & ($10^{-3}$) & ($^{\circ}$) & ($10^{-2}$) \\
        \midrule 
        PMT & - &  \textbf{15.80} & \textbf{5.55} & 71.91 & \textbf{14.31} \\
        PMT & PMT & 16.01 & 6.01 & \textbf{70.13} & 15.70 \\
        \bottomrule
	\end{tabular}
	}
    \caption{
		Ablation study on the feature matcher in PMTR~\cite{lee2024pmtr}.
	}
\label{table:ambiguous_pmtr}
\end{table}

%% file: figures/suppl_heatmap.tex

\begin{figure}[t]
\begin{center}
\includegraphics[width=\linewidth]{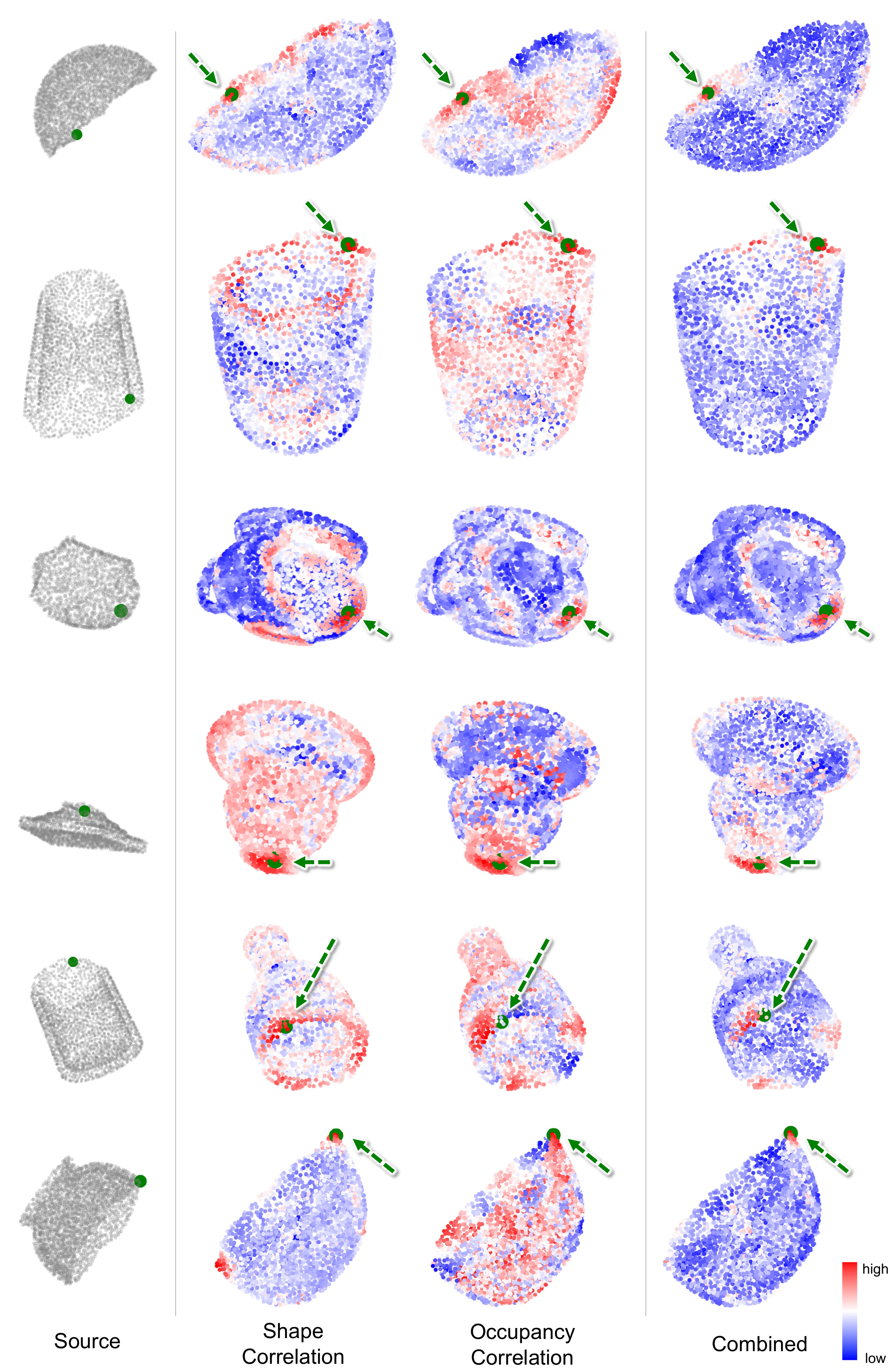}
\vspace{-3mm}
\caption{\textbf{Additional visualization of correlation distribution.} 
A green dot (\textcolor[HTML]{22741C}{$\bullet$}) on the left point cloud marks the source's $i$-th point, with corresponding true match points marked with green dots and arrows.
Point colors represent correlation score magnitudes for the $i$-th point's similarity to each target point, with \textcolor{red}{red} and \textcolor{blue}{blue} indicating high and low correlation scores, respectively.}
\label{fig:suppl_heatmap}
\end{center}
\vspace{-7mm}
\end{figure}

%% file: figures/suppl_orientation.tex

\begin{figure*}[ht]
\begin{center}
\includegraphics[width=.95\linewidth]{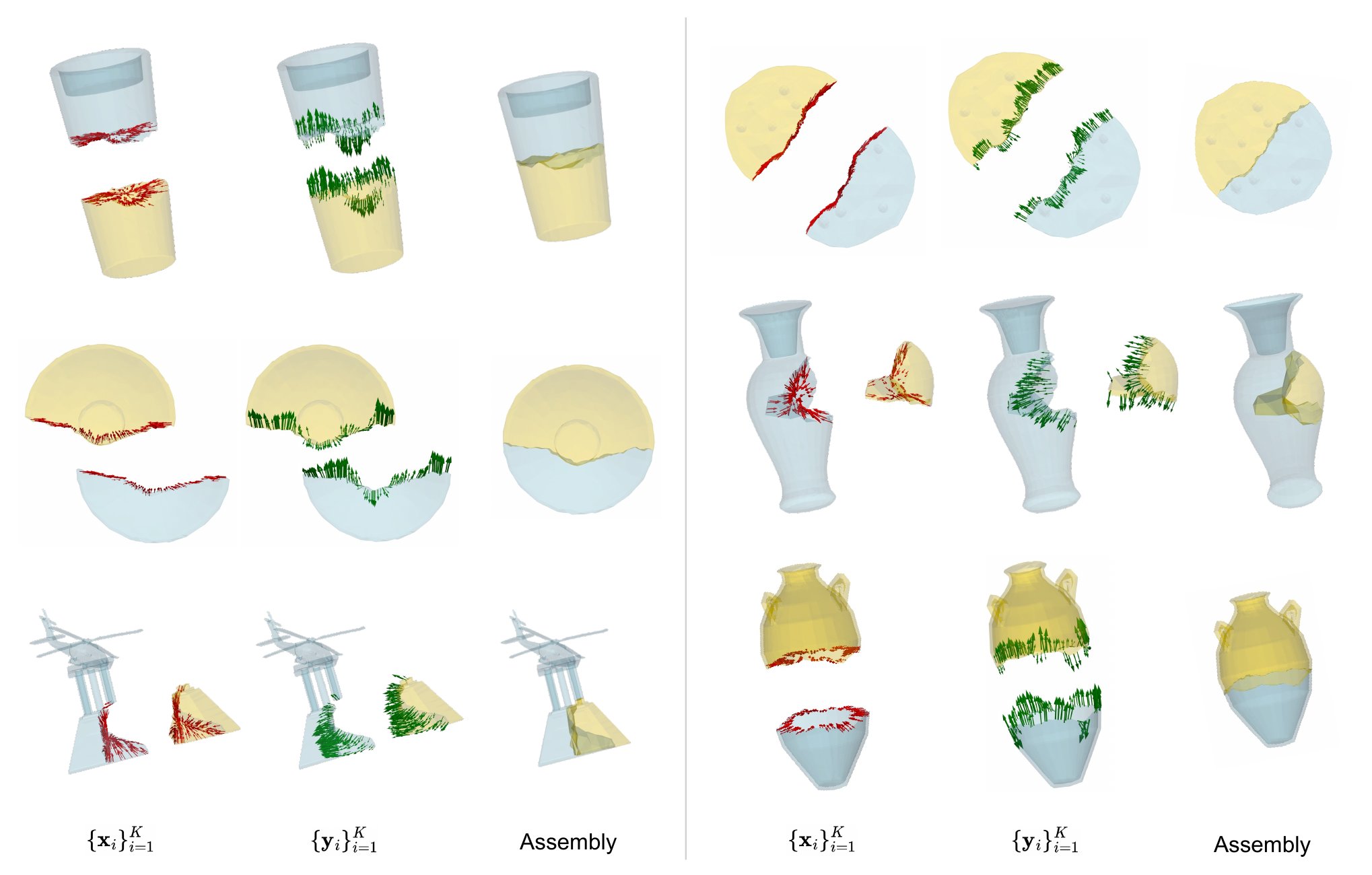}
\vspace{-2mm}
\caption{\textbf{Additional visualization of learned orientations.} We visualize learned orientations (in $\mathbb{R}^{3}$) of $\{\mathbf{x}_i\}_{i \in \mathcal{I}}$ (left, \textcolor[HTML]{CC0000}{red arrows}) and $\{\mathbf{y}_i\}_{i \in \mathcal{I}}$ (middle, \textcolor[HTML]{22741C}{green arrows}). The assembly results are shown on the right.}
\label{fig:suppl_orientation}
\end{center}
\vspace{-7mm}
\end{figure*}

%% file: sections/suppl/2_model_details.tex

\section{Additional Network Details}
\label{sec:additional_method_detail}
In this section, we provide the details of the network components that were omitted in the main paper for brevity.

\input{figures/suppl_backbone}

\subsection{Equivariant feature extractor}
For our backbone network $f_{\text{VNN}}$, we adopt Vector Neuron Network (VNN)~\cite{deng2021vector}, which represents neurons as 3D vectors, \ie, $(\mathbf{F}_{\text{eqv}})_{i,j} \in \mathbb{R}^{3}$ for all $i,j$.
This representation enables the network to handle $\text{SO(3)}$ transformations directly, preserving consistent local feature orientations.
Specifically, for any given input feature $\mathbf{F}$, each $i$-th layer $f_{\text{VNN}}^{i}$ of the network satisfies the following property:
$f_{\text{VNN}}^{i}(\mathbf{F}\mathbf{R})=f_{\text{VNN}}^{i}(\mathbf{F})\mathbf{R}$ where $\mathbf{R} \in \text{SO(3)}$~\cite{deng2021vector}.
To enhance the general context learning capabilities of the network, we modify the original VN-DGCNN~\cite{wang2019dynamic, deng2021vector} architecture to broaden the receptive field of the features $\mathbf{F}_{\text{eqv}}$ by redesigning the network into a U-shaped architecture, as illustrated in Fig.~\ref{fig:backbone}.
For downsampling and upsampling of the features in the process, we utilize the TransitionDown and TransitionUp modules, similar to the approach in~\cite{zhao2021point}.
This modification allows for an efficient contextual feature extraction while preserving the rotational equivariance property.

\subsection{Orientation hypothesizer}
The backbone output $\mathbf{F}_{\text{eqv}} \in \mathbb{R}^{K \times D \times 3}$ is processed through an orientation hypothesizer $f_{\text{hyp}}$ to provide orientations such that $f_{\text{hyp}}(\mathbf{F}_{\text{eqv}}) = \mathbf{F}_{\text{d}} \in \mathbb{R}^{K \times 3 \times 3}$. 
The hypothesizer contains a VN-Linear~\cite{deng2021vector} layer that reduces the channel dimension of the input features from $D$ to $2$, producing two vectors of size $\mathbb{R}^{K \times 2 \times 3}$, where these two vectors, for each point, represent candidate orientations for the $x$-axis and $y$-axis, respectively.
To ensure these vectors form a valid orientation basis, we apply the Gram-Schmidt process:
First, we compute the 2D plane spanned by the two vectors and adjust the $y$-axis vector to ensure it is orthogonal to the other $x$-axis vector within this plane.
Next, we calculate a third vector orthogonal to the 2D plane, resulting in a complete orthonormal basis.
Finally, we apply L2 normalization to these vectors, ensuring that each $(\mathbf{F}_{\text{d}})_i \in \text{SO(3)}$ for all $i$, representing a valid 3D rotation matrix for each point.
Note that this Gram-Schmidt process is rotation-equivariant, \ie, $f_{\text{hyp}}(\mathbf{F}_{\text{eqv}}\mathbf{R}) = f_{\text{hyp}}(\mathbf{F}_{\text{eqv}})\mathbf{R}$ for any $\mathbf{R} \in \text{SO(3)}$, as discussed by Luo~\etal~\cite{luo2022equivariant}.

\subsection{Invariant feature computation}
Given the equivariant network $f_{\text{VNN}}$ which satisfies $f_{\text{VNN}}(\mathbf{X}\mathbf{R}) = f_{\text{VNN}}(\mathbf{X})\mathbf{R}$ for any rotations $\mathbf{R} \in \text{SO(3)}$ and input points $\mathbf{X} \in \mathbb{R}^{K \times 3}$, along with the hypothesizer $f_{\text{hyp}}$, we aim to define a function $f_{\text{inv}}$ that provides invariant features $\mathbf{F}_{\text{inv}} \in \mathbb{R}^{K \times 3D}$, satisfying the following property:
\begin{align}
    f_{\text{inv}}(\mathbf{X}) = f_{\text{inv}}(\mathbf{X}\mathbf{R}) = \mathbf{F}_{\text{inv}},
\end{align}
for any rotation matrix $\mathbf{R}$.
To achieve this, we define $f_{\text{inv}}$ as the dot product between the equivariant features and the hypothesized orientations:
\begin{align}
    f_{\text{inv}}(\mathbf{X}_i) = (\mathbf{F}_{\text{eqv}})_{i} \cdot (\mathbf{F}_{\text{d}})_{i}^{\top},
\end{align}
for all $i \in \{1, \dots, K\}$, where $(\mathbf{F}_{\text{eqv}})_{i}$ represents the equivariant features output by $f_{\text{VNN}}$ and $(\mathbf{F}_{\text{d}})_{i}$ represents the hypothesized orientations output by $f_{\text{hyp}}$. The invariance property of $f_{\text{inv}}$ can be verified through the following proof:
\begin{align}
    f_{\text{inv}}(\mathbf{X}_i\mathbf{R}) &= f_{\text{VNN}}(\mathbf{X}_i\mathbf{R}) \cdot (f_{\text{hyp}}(f_{\text{VNN}}(\mathbf{X}_i\mathbf{R})))^{\top} \\ \nonumber
    &= (f_{\text{VNN}}(\mathbf{X}_i) \mathbf{R}) \cdot (f_{\text{hyp}}(f_{\text{VNN}}(\mathbf{X}_i))\mathbf{R})^{\top} \\ \nonumber
    &= f_{\text{VNN}}(\mathbf{X}_i) \mathbf{R} \mathbf{R}^{\top} (f_{\text{hyp}}(f_{\text{VNN}}(\mathbf{X}_i)))^{\top} \\ \nonumber
    &= f_{\text{VNN}}(\mathbf{X}_i) (f_{\text{hyp}}(f_{\text{VNN}}(\mathbf{X}_i)))^{\top} \\ \nonumber
    &= (\mathbf{F}_{\text{eqv}})_i \cdot (\mathbf{F}_{\text{d}})_i^{\top} \\ \nonumber
    &= f_{\text{inv}}(\mathbf{X}_i). \nonumber
\end{align}
Thus, $f_{\text{inv}}$ is provably invariant to rotations, making $\mathbf{F}_{\text{inv}}$ suitable for the subsequent tasks requiring rotational invariance, such as shape and occupancy matching.

\input{tables/suppl_1_matcher}
\subsection{Shape and occupancy matcher}
In Tab.~\ref{table:suppl_matcher}, we tabularize the components of each layer for shape and occupancy matchers.
Each matcher consists of a three-layer MLP, where each layer includes a linear transformation followed by normalization and activation.
The key difference between shape and occupancy matchers lies in the activation function used in the final layer:
LeakyReLU is used for the shape matcher to allow the shape correlation matrix $\mathbf{C}_{\text{s}}$ to have a wider range of values, capturing large variations in shape similarity.
In contrast, Tanh is employed for the occupancy descriptors to constrain extreme activations, ensuring that the occupancy correlation matrix $\mathbf{C}_{\text{o}}$ avoids overemphasizing noisy or outlier regions, particularly those unrelated to occupancy learning, such as non-mating surfaces.

Empirical observations, as shown in Fig.~\ref{fig:suppl_heatmap}, indicate that outliers occur more frequently in occupancy correlations compared to shape correlations, with large occupancy scores being more uniformly distributed across surfaces, whereas shape scores are more localized and structured.
Allowing large variations in $\mathbf{C}_{\text{s}}$, therefore, ensures that dominant shape features are captured (introducing local ambiguity), while controlling $\mathbf{C}_{\text{o}}$ prevents outliers or irrelevant regions from skewing the overall correlation (resolving the ambiguity), resulting in more reliable correlations $\mathbf{C}$.

\input{figures/attention}

When generating shape and occupancy descriptors, we dynamically adjust the importance of feature channels using soft-channel attention, inspired by SENet~\cite{hu2018squeeze}. 
As illustrated in Fig.~\ref{fig:attention}, we first concatenate the pair of invariant features along the spatial dimension and apply average-pooling and max-pooling.
The pooled outputs are passed to two shared MLPs followed by a sigmoid activation to compute channel-wise statistics, which serve as the soft-attention values for the shape and occupancy descriptors.
The output $\mathbf{A}$ is divided by two along the channel dimension to produce $\mathbf{A}_{\text{s}}$ and $\mathbf{A}_{\text{o}}$, the soft attention weights for the shape and occupancy descriptors, respectively, each of which weights the feature channels to enhance relevant information for each descriptor as in~\cite{hu2018squeeze}.

\subsection{Details on training objectives}

\smallbreak\noindent\textbf{Circle loss~\cite{sun2020circle}.}
The definition of $\mathcal{L}_{\text{circle}}$ is formulated as 
\begin{align}
    \underset{i \sim \mathcal{I}}{\mathbb{E}} \left[ \log \left( \sum_{j \in \mathcal{E}_{\text{p}}(i)} e^{\alpha_{ij}(\phi^{i,j} - \Delta_{\text{p}})} \cdot \sum_{k \in \mathcal{E}_{\text{n}}(i)} e^{\beta_{ik}(\Delta_{\text{n}} - \phi^{i,k})} \right) \right],
\end{align}
where the positive and negative weights are defined as
$\alpha_{ij}=\gamma[\phi^{i,j}-\Delta_\text{p}]_{+}$ and $\beta_{ik}=\gamma[\Delta_\text{n}-\phi^{i,k}]_{+}$,
where $\gamma$ controls the sharpness of the exponential terms, amplifying or reducing the emphasis on outliers and enabling more stable training. 
And $\phi^{i,j}$ denotes the similarity (or dissimilarity) between a given feature pair.
In our implementation, the circle loss uses margin parameters of $\Delta_{\text{p}} = 0.1$ and $\Delta_{\text{n}} = 1.4$, with $\gamma = 24$.

\smallbreak\noindent\textbf{Point matching loss~\cite{sarlin2020superglue}.}
We now detail the point matching loss $\mathcal{L}_{\text{p}}$ which is jointly used alongside the combinative matching objectives.
The point matching loss is a negative log-likelihood loss on predicted match probabilities $\mathbf{Z} \in \mathbb{R}^{(N+1) \times (M+1)}$, the dual-normalized assignment matrix with dustbin, the output from the Optimal Transport layer~\cite{sarlin2020superglue}. 
Given a set of ground-truth correspondences $\mathcal{C}$, and the sets of unmatched points, $(\mathcal{I}^{\text{P}})' = \{x: x \in [N] \wedge x \notin \mathcal{I}^{\text{P}}\}~\text{and}~(\mathcal{I}^{\text{Q}})' = \{x: x \in [M] \wedge x \notin \mathcal{I}^{\text{Q}}\}$ where $[K] = \{1, \dots, K\}$, the point matching loss is defined as
\begin{align}
    \mathcal{L}_{\text{p}} = &- \sum_{(i,j)\in \mathcal{C}}\log{\mathbf{Z}_{i,j}} \nonumber \\
    &- \sum_{i \in ({\mathcal{I}^{\text{P}}})'} \log{\mathbf{Z}_{i,M+1}} - \sum_{j \in ({\mathcal{I}^{\text{Q}}})'} \log{\mathbf{Z}_{N+1, j}},
\end{align}
which enforces the predicted match probabilities to align closely with the ground-truth matches, ensuring accurate point-to-point correspondence.

%% file: figures/suppl_backbone.tex

\begin{figure*}[!t]
\begin{center}
\includegraphics[width=1\linewidth]{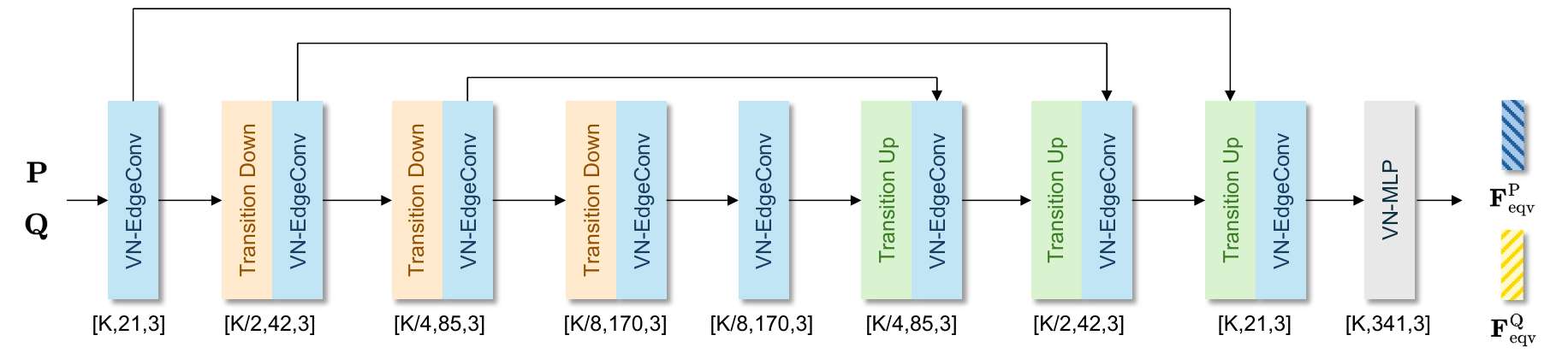}
\vspace{-6mm}
\caption{Overall framework of our U-shaped equivariant feature extractor.}
\label{fig:backbone}
\end{center}
\vspace{-7mm}
\end{figure*}

%% file: tables/suppl_1_matcher.tex

\begin{table}[!t]
    \centering
    \resizebox{.85\linewidth}{!}{
	\begin{tabular}{c|cc}
        \toprule
        Layer & Shape Matcher & Occupancy Matcher \\
        \midrule
        \multirow{3}{*}{1} & Conv1D(1023 $\to$ 512) & Conv1D(1023 $\to$ 512) \\
        & InstanceNorm(512) & InstanceNorm(512) \\
        & LeakyReLU & LeakyReLU \\
        \midrule
        \multirow{3}{*}{2} & Conv1D(512 $\to$ 512) & Conv1D(512 $\to$ 512) \\
        & InstanceNorm(512) & InstanceNorm(512) \\
        & LeakyReLU & LeakyReLU \\
        \midrule
        \multirow{3}{*}{3} & Conv1D(512 $\to$ 512) & Conv1D(512 $\to$ 512) \\
        & InstanceNorm(512) & InstanceNorm(512) \\
        & LeakyReLU & Tanh \\
        \bottomrule
	\end{tabular}
    }
    \caption{Components of the shape and occupancy matchers.}
    \label{table:suppl_matcher}
\end{table}

%% file: figures/attention.tex

\begin{figure}[h]
\begin{center}
\includegraphics[width=\linewidth]{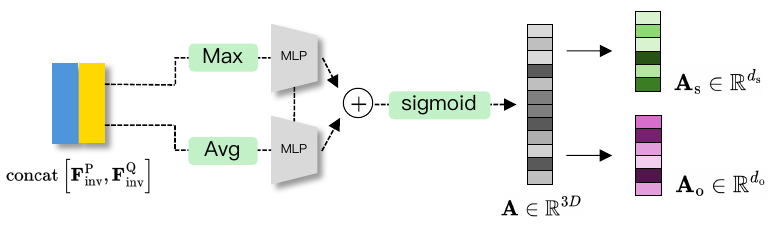}
\vspace{-3mm}
\caption{Pipeline for soft-attention generation.}
\label{fig:attention}
\end{center}
\vskip -0.2in
\end{figure}

%% file: sections/suppl/3_exp_setup.tex

\section{Details on Training and Evaluation Setup}
\label{sec:additional_setup}

\subsection{Hyperparameter setup}
To determine positive matches between mating surfaces, the distance threshold was set to $\tau=0.018$.
For the circle loss~\cite{sun2020circle}, we used margin hyperparameters $\Delta_p=0.1$ and $\Delta_n=1.4$, along with a scale factor $\gamma=24$.
The channel sizes for the equivariant feature embedding, shape descriptor, and occupancy descriptor are set to $D=341$, $d_{\text{s}}=512$, and $d_{\text{o}}=512$, respectively.
We use the normalization constant $Z=\sqrt{512}$ to construct the cost matrix $\mathbf{C}$.

\subsection{Multi-part assembly details}
Following the approach of Lee~\etal~\cite{lee2024pmtr}, we extend our pairwise matching framework \oursabb{} to handle multi-part assembly in a consistent manner.
Specifically, our method first learns \textbf{\textit{local pairwise compatibilities}} between part pairs through pairwise matching, and then estimates globally consistent poses via \textbf{\textit{pose graph optimization}}.

\smallbreak
\noindent\textbf{Learning Pairwise Compatibility.}  
We construct 2-part training pairs from all objects in the Breaking Bad dataset~\cite{sellan2022breakingbad}, where each object consists of 2 to 20 parts. 
Specifically, for each training sample, we randomly select a \textit{source} part and choose as its \textit{target} part the one that shares the largest ground-truth mating surface area with it.
We train the network for 350 epochs on the \texttt{everyday} subset and 300 epochs on the \texttt{artifact} subset, using the same model configurations (\eg, hyperparameters) as in the pairwise setup.
This training is intended to allow the model to learn local pairwise compatibilities, which are subsequently utilized during global optimization in the multi-part setting.

\begin{wrapfigure}{r}{0.20\linewidth}
  \centering
  \vspace{-9mm}
  \hspace{-8mm}
  \includegraphics[width=1.3\linewidth]{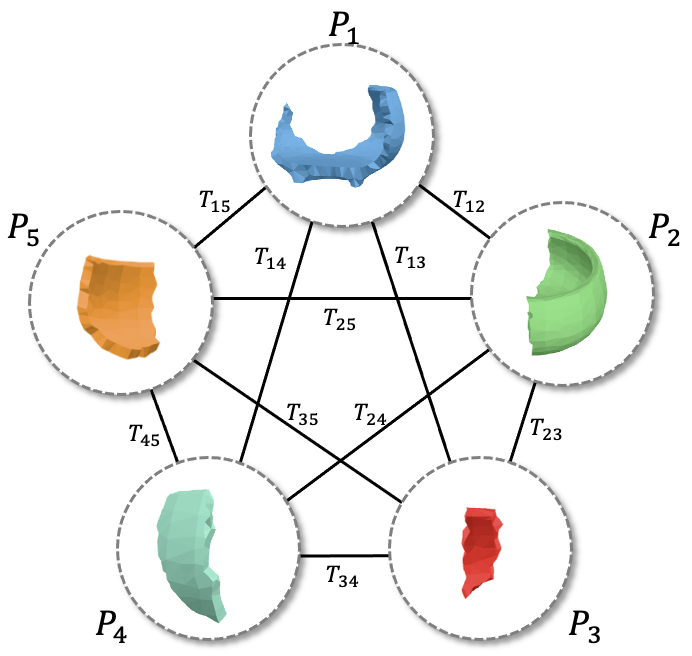}
  \vspace{-2mm}
  \captionsetup{font=footnotesize, justification=raggedright, singlelinecheck=false, margin=0pt}
  \hspace{-2mm}
  \caption{Pose graph example.}
  \vspace{-8mm}
  \label{fig:rebuttal_pgo}
\end{wrapfigure}

\smallbreak
\noindent\textbf{Inference \& Pose-Graph Optimization.} 
Given an $N$-part object at the inference time, we first predict relative poses for all $\binom{N}{2}$ part pairs to construct a fully connected pose graph, as illustrated in Fig.~\ref{fig:rebuttal_pgo}.

In the \textit{pose graph}, each part $P_i$ (with $i = 1, \dots, N$) becomes a node, and each weighted edge carries a \textit{pairwise relative pose} $T_{ij} = \{\mathbf{R}_{ij}, \mathbf{t}_{ij}\}$ along with its information matrix $\mathbf{I}_{ij} = \left(\sum_{p=1}^{|P_i|} \sum_{q=1}^{|P_j|} \exp(\mathbf{Z}^{(ij)}_{p,q})\right)^{-1} \cdot I_6$, where $\mathbf{Z}^{(ij)} \in \mathbb{R}^{|P_i| \times |P_j|}$ denotes the soft assignment matrix between parts $P_i$ and $P_j$, obtained via Optimal Transport~\cite{sarlin2020superglue}.

To reduce noise and eliminate unreliable matches, we prune all outgoing edges except the one with the highest matchability score for each node.
We then recover global poses $(\tilde{\mathbf{R}}_i, \tilde{\mathbf{t}}_i)$ via Shonan Averaging~\cite{dellaert2020shonan}, with the largest part set as an anchor, similar to~\cite{lu2023jigsaw, lee2024pmtr}.

\clearpage

%% file: figures/suppl_pairwise_qual.tex

\begin{figure*}[!t]
\begin{center}
\includegraphics[width=\linewidth]{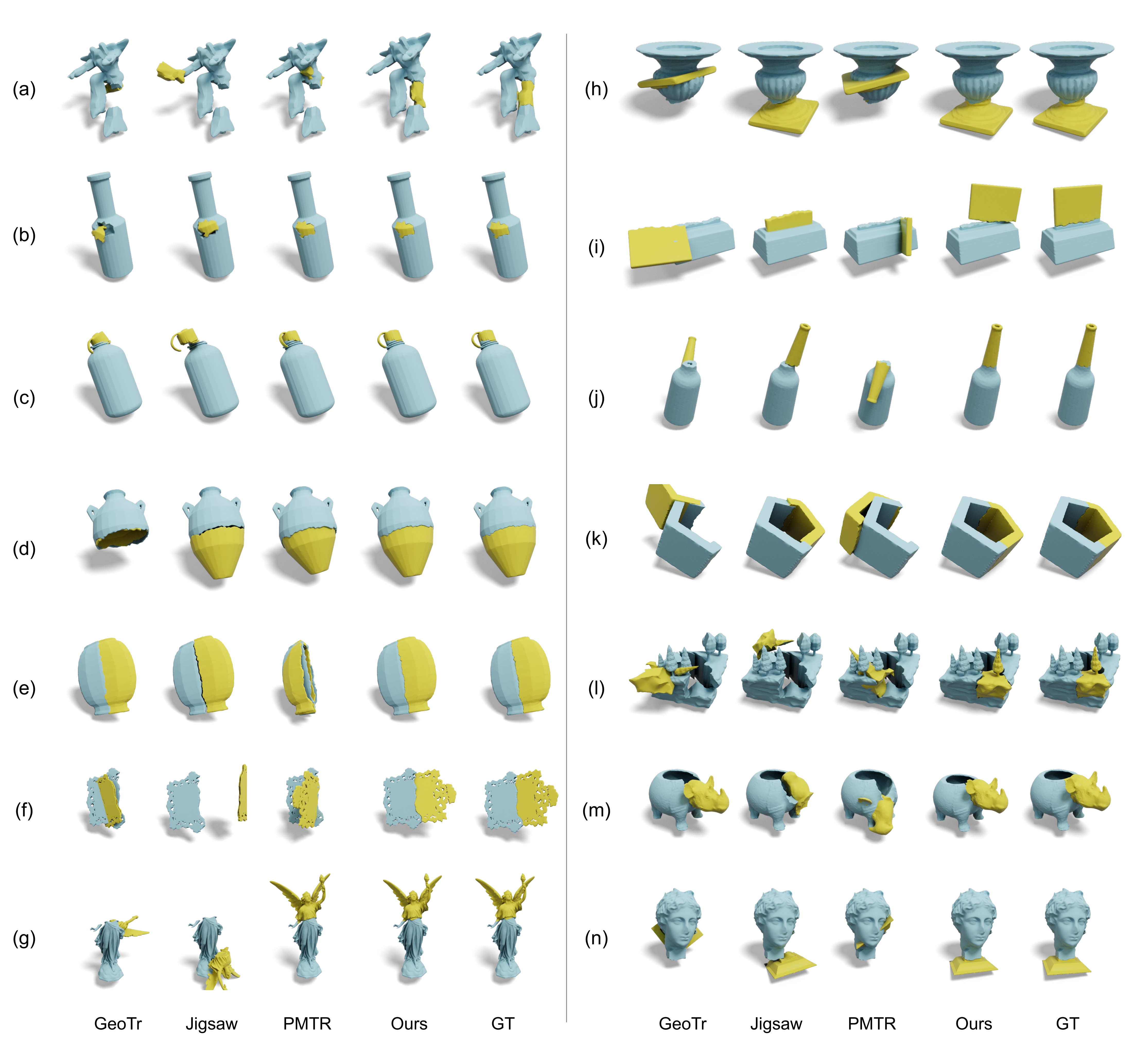}
\vspace{-6mm}
\caption{Additional qualitative comparison for pairwise shape assembly.}
\label{fig:suppl_pairwise_qual}
\end{center}
\vspace{-7mm}
\end{figure*}

%% file: figures/suppl_mpa_qual.tex

\begin{figure*}[!t]
\begin{center}
\includegraphics[width=\linewidth]{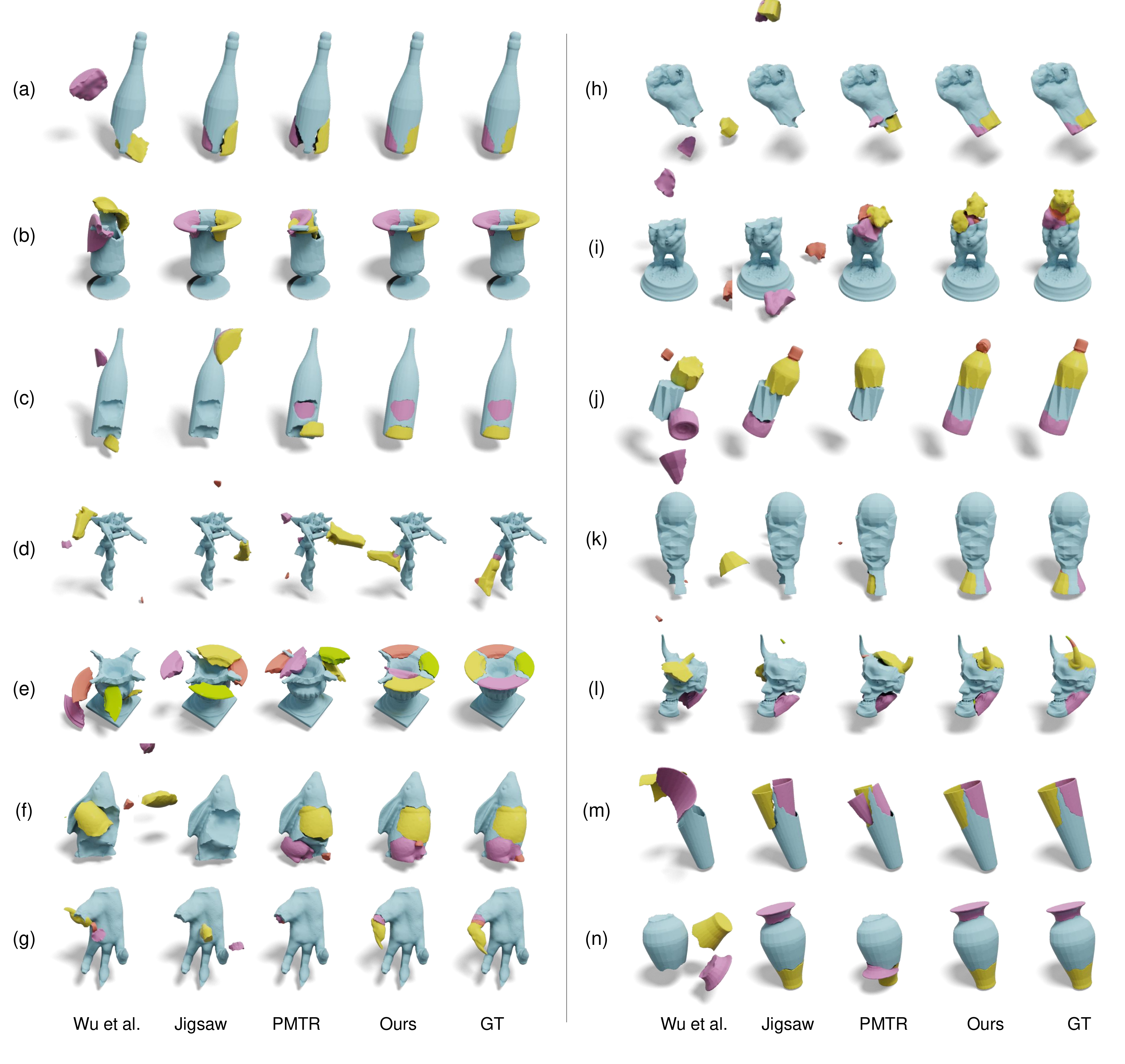}
\vspace{-6mm}
\caption{Additional qualitative comparison for multi-part assembly.}
\label{fig:suppl_mpa_qual}
\end{center}
\vspace{-7mm}
\end{figure*}